\documentclass[11pt,usenames,dvipsnames,a4paper]{article}
\usepackage{acl2023}
\usepackage{times}
\usepackage{latexsym}

\usepackage{times}
\usepackage{latexsym}
\usepackage{tikz}
\usepackage{amsmath,amssymb}
\usepackage{xcolor}
\usepackage{colortbl}
\usepackage{tabularx}
\usepackage{subfig}
\usetikzlibrary{fit,positioning}
\usepackage{mathtools}
\usepackage{microtype}
\usepackage{multirow}
\usepackage{hhline}
\usepackage{array}
\usepackage{booktabs}
\usepackage{amsmath}
\usepackage{makecell}
\usepackage{pbox}
\usepackage{cellspace}
\usepackage{subfig}
\usepackage{balance}
\usepackage{textcomp}

\usepackage{soul}
\usepackage{caption}
\usepackage{newfloat}
\usepackage{appendix}
\usepackage{graphicx}
\usepackage{scalerel,xparse}

\NewDocumentCommand\circlered{}{
	\includegraphics[scale=0.06]{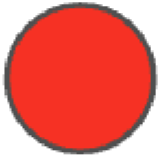}
}
\NewDocumentCommand\circlepeach{}{
	\includegraphics[scale=0.06]{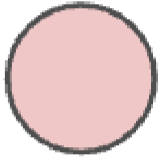}
}
\NewDocumentCommand\circlebeige{}{
	\includegraphics[scale=0.06]{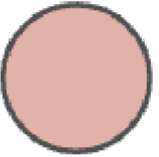}
}
\NewDocumentCommand\squareblue{}{
	\includegraphics[scale=0.06]{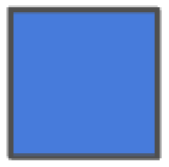}
}
\NewDocumentCommand\squarepurple{}{
	\includegraphics[scale=0.06]{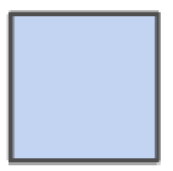}
}
\NewDocumentCommand\squareteel{}{
	\includegraphics[scale=0.06]{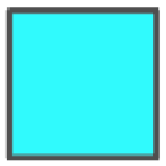}
}

\DeclareFloatingEnvironment[listname=Box,name=Box]{story}

\expandafter\def\expandafter\UrlBreaks\expandafter{\UrlBreaks%  save the current one
	\do\a\do\b\do\c\do\d\do\e\do\f\do\g\do\h\do\i\do\j%
	\do\k\do\l\do\m\do\n\do\o\do\p\do\q\do\r\do\s\do\t%
	\do\u\do\v\do\w\do\x\do\y\do\z\do\A\do\B\do\C\do\D%
	\do\E\do\F\do\G\do\H\do\I\do\J\do\K\do\L\do\M\do\N%
	\do\O\do\P\do\Q\do\R\do\S\do\T\do\U\do\V\do\W\do\X%
	\do\Y\do\Z}

\definecolor{Statement}{RGB}{188,189,34}
\definecolor{purple}{RGB}{230,210,219}
\definecolor{Quote}{RGB}{44,160,44}
\definecolor{Order}{RGB}{23,190,207}
\definecolor{azure}{RGB}{211, 223, 226}
\definecolor{bronze}{rgb}{0.8, 0.5, 0.2}
\definecolor{darkpastelpurple}{rgb}{0.59, 0.44, 0.84}
\definecolor{darktangerine}{rgb}{1.0, 0.66, 0.07}
\usepackage{array,multirow}
\definecolor{deepchampagne}{rgb}{0.98, 0.84, 0.65}
\definecolor{dollarbill}{RGB}{220, 233, 213}
\definecolor{lemon}{rgb}{1.0, 0.97, 0.0}
\definecolor{lightsalmon}{rgb}{1.0, 0.63, 0.48}
\definecolor{lavenderblue}{rgb}{0.8, 0.8, 1.0}

\DeclareRobustCommand{\hlpurple}[1]{{\sethlcolor{purple}\hl{#1}}}
\DeclareRobustCommand{\hlgreen}[1]{{\sethlcolor{dollarbill}\hl{#1}}}
\DeclareRobustCommand{\hlazure}[1]{{\sethlcolor{azure}\hl{#1}}}

\makeatletter
\def\gcmidrule{\arrayrulecolor{lightgray}% Switch colour to lightgray
	\noalign{\ifnum0=`}\fi
	\@ifnextchar[{\@gcmidrule}{\@gcmidrule[\cmidrulewidth]}}
\def\@gcmidrule[#1]{\@ifnextchar({\@@gcmidrule[#1]}{\@@gcmidrule[#1]()}}
\def\@@gcmidrule[#1](#2)#3{\@@@gcmidrule[#3]{#1}{#2}}
\def\@@@gcmidrule[#1-#2]#3#4{\global\@cmidla#1\relax
	\global\advance\@cmidla\m@ne
	\ifnum\@cmidla>0\global\let\@gtempa\@cmidrulea\else
	\global\let\@gtempa\@cmidruleb\fi
	\global\@cmidlb#2\relax
	\global\advance\@cmidlb-\@cmidla
	\global\@thisrulewidth=#3
	\@setrulekerning{#4}
	\ifnum\@lastruleclass=\z@\vskip \aboverulesep\fi
	\ifnum0=`{\fi}\@gtempa
	\noalign{\ifnum0=`}\fi\futurenonspacelet\@tempa\@xgcmidrule}
\def\@xgcmidrule{%
	\ifx\@tempa\gcmidrule
	\vskip-\@thisrulewidth
	\global\@lastruleclass=\@ne
	\else \ifx\@tempa\morecmidrules
	\vskip \cmidrulesep
	\global\@lastruleclass=\@ne\else
	\vskip \belowrulesep
	\global\@lastruleclass=\z@
	\fi\fi
	\ifnum0=`{\fi}
	\arrayrulecolor{black}}% Switch colour back to black
\makeatother

\title{Explaining Mixtures of Sources in News Articles}

\author{Alexander Spangher\textsuperscript{1}, James Youn\textsuperscript{2}, Matt DeButts\textsuperscript{3},\\ \textbf{Nanyun Peng\textsuperscript{2}}, \textbf{Emilio Ferrara\textsuperscript{1}}, \textbf{Jonathan May\textsuperscript{1}} \\
	\textsuperscript{1}University of Southern California \\
	\textsuperscript{2}University of California, Los Angeles, 
	\\ 
	\textsuperscript{3} Stanford University  \\ 
	\texttt{spangher@usc.edu}
}

\date{}

\begin{document}
	% \doparttoc % Tell to minitoc to generate a toc for the parts
	% \faketableofcontents % Run a fake tableofcontents command for the partocs
	
	% \part{} % Start the document part
	% \parttoc % Insert the document TOC

	\maketitle
	
	\begin{abstract}
		% Writers often need to draw information from multiple different sources before writing. Although the choice of sources is key to the overall synthesis of information in the written output, little is understood about why sources are chosen together. 
		% Shedding light on such questions could help us \textit{analyze} existing writing, as well \textit{aid} future writing by performing more useful retrieval.
		% Decisions about which sources to draw information from in news articles is a key planning step in the writing process. Explaining these decisions could help us about why different sources are chosen could help analyze 
		%Decisions about which sources to draw informational from is a key planning step in the writing of news articles, yet, little is understood about why certain sets of sources are included together in articles. % Are they chosen because they are oppositional? Complementary? 
		% In this work, we take the view that well-written news articles group sources together in order to put different groups in dialogue to reach a richer telling of the story. 
		
		% In this work, we compare explanations for the choice of sources in writing. 
		% : \textit{stance}, \textit{argumentation}, \textit{discourse}, \textit{NLI}, \textit{retrieval-channel}
		% Is the choice of sources driven by a desire to include ? 
		%  (i.e. \textit{imitation learning with unobserved action sequences})
		% 
		% first understand how humans create plans. 
		Human writers plan, \textit{then} write \cite{yao2019plan}. 
		For large language models (LLMs) to play a role in longer-form article generation, we must understand the planning steps humans make before writing. 
		% 
		% Although large lanuage models (LLMs) are being explored in contexts which require planning, they so far have struggled to take basic steps
		% 
		We explore one kind of planning, source-selection in news, as a case-study for evaluating plans in long-form generation. We ask: why do \textit{specific} stories call for \textit{specific} kinds of sources? We imagine a generative process for story writing where a source-selection schema is first selected by a journalist, and then sources are chosen based on categories in that schema.
		Learning the article's \textit{plan} means predicting the schema initially chosen by the journalist.
		% with the goal of helping journalists \textit{plan}, given a story idea, the set of sources they will select. 
		Working with professional journalists, we adapt five existing schemata and introduce three new ones to describe journalistic plans for the inclusion of sources in documents.  Then, inspired by Bayesian latent-variable modeling, we develop metrics to select the most likely plan, or schema, underlying a story, which we use to compare schemata. We find that two schemata: \textit{stance} \cite{hardalov2021cross} and \textit{social affiliation} best explain source plans in most documents. %reduces perplexity the most overall (10\% below baseline)%is most important for describing a source's contribution to news articles
		However, other schemata like \textit{textual entailment} explain source plans in factually rich topics like ``Science''. %We compare against three unsupervised latent variable models we design. 
		Finally, we find we can predict the most suitable schema given just the article's headline with reasonable accuracy. We see this as an important case-study for human planning, and provides a framework and approach for evaluating other kinds of plans.
		We release a corpora, \textit{NewsSources}, with annotations for 4M articles.
		
		%In this work, we answer this question with a very simple conceptual model: all sources are divided into \textit{canonical groups}; for each piece of writing, sources from different groups are sampled; different story topics call for different combinations of groups. 
		
		%We realize this conceptualization \textit{in the news domain} by developing a novel typology of sources, with which we categorize over 4,500 sources from over 520 news articles.\footnote{We release this dataset, which we call \textit{NewsSources}}. This typology is simple to learn: BERT-based classification models can score new sources with $>.8$ f1.
		
		% We show that this conceptual model is (1) a useful descriptive and analytical tool by creating a large silver-standard corpus of over YY million sources from 1 million news articles spanning 30 years, with which we examine economic, political and editorial changes in news media over this time period. And (2) a useful aid for planning before writing. We show that, given a simple description of a story topic, large language models can predict the different types of sources required for stories with high retrieval accuracy. We anticipate this work can open the door to future work in narrative planning, source recommendation enginges and media studies.
	\end{abstract}
	
	\section{Introduction}
	
	\begin{table}[t!]
		% \small 
		% \begin{tabular}{p{12.7cm}p{2.6cm}}
			\centering
			\begin{tabular}{p{7cm}} 
				\toprule
				\textbf{Headline: NJ Schools Teach Climate Change at all Grade Levels} \\
				\midrule
				\hspace {4mm} \textbf{Michelle Liwacz} asked her first graders: what can penguins do to adapt to a warming Earth?  \hspace{.23cm} $\leftarrow$  {\small \textit{potential labels:}} \hlgreen{Academic}, \hlpurple{Neutral} \\
				\hspace{4mm} \textbf{Gabi}, 7, said a few could live inside her fridge. \hspace{.05cm} $\leftarrow$ {\small \textit{potential labels:}} \hlgreen{Unaffiliated}, \hlpurple{Neutral} \\
				\hspace{4mm} \textbf{Tammy Murphy}, wife Governor Murphy, said climate change education was vital to help students. \hspace{.01cm} $\leftarrow$ {\small \textit{poten. labels:}} \hlgreen{Government}, \hlpurple{Agree} \\
				\hspace{4mm} \textbf{Critics} said young kids shouldn't learn disputed science.
				\hspace{.1cm} 
				$\leftarrow$ {\small \textit{labels:}} 
				\hlgreen{Unaffiliated}, \hlpurple{Refute} \\
				\hspace{4mm} A \textbf{poll} found that 70 percent of state residents supported climate change being taught at schools. \hfill $\leftarrow$ {\small \textit{potential labels:}} \hlgreen{Media}, \hlpurple{Agree}  \\
				\bottomrule
			\end{tabular}
			\caption{Informational sources synthesized in a single news article. \textit{How would we choose sources to tell this story?} We show two different explanations, given by two competing schemata: \hlgreen{affiliation} and \hlpurple{stance}. Our central questions: (1) \textit{Which schema best explains the sources used in this story?} (2) \textit{Can we predict, given a topic sentence, which schema to use?}}
			\label{tbl:intro-example-article}
		\end{table}
		
		%  conflicting viewpoints, or to reflect the perspectives of varied groups? 
		%Journalists might make such choices for many different reasons\footnote{Many explanations might exist for why sources are chosen together. In the example given, centralized newsroom structures \cite{christin2020metrics} guide the planning; repetitive reporting patterns (i.e. ``beats'') \cite{magin2019beat} or journalistic ideals \cite{dzur2002public, conrad1999uses} can also inform a plan.}. 
		%Still others might call for a dialogue between sources with different \textit{affiliations} (e.g. ``government'', ``academic'', ``corporate'').
		
		\begin{figure*}[t]
			\includegraphics[width=\linewidth]{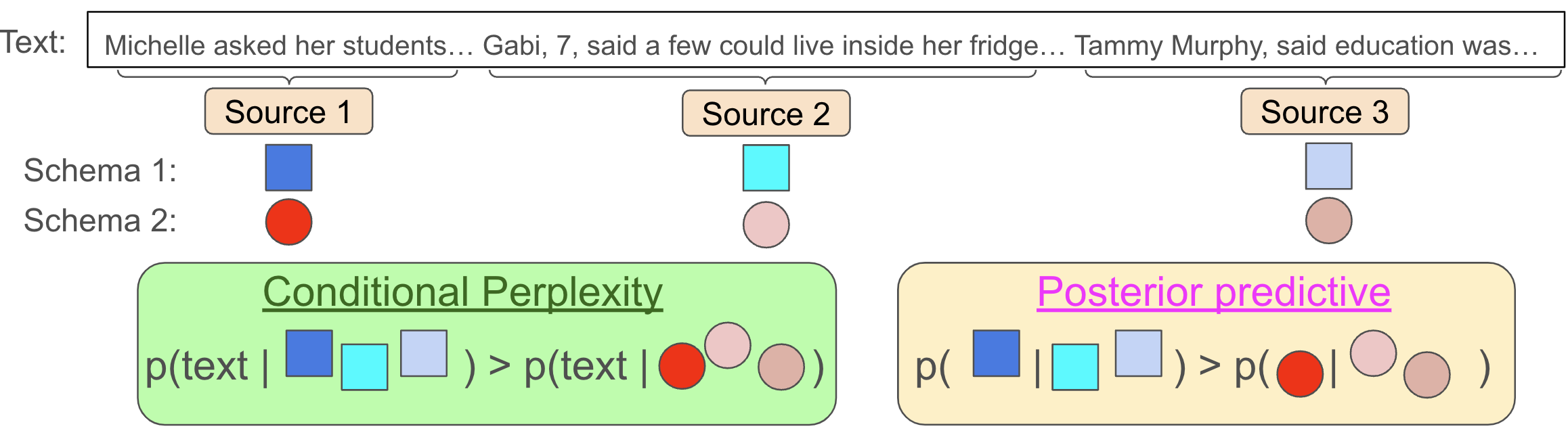}
			\caption{We seek to infer unobserved \textit{plans}, or schemata, in natural data, focusing on one scenario: source-selection made by human journalists during news writing. Although the \textit{reasons} why sources are chosen are unobservable, we show that one explanation (in the diagram, represented by \textit{squares}: \{\squareblue, \squareteel, \squarepurple\}), is preferred over another (represented by \textit{circles}: \{\circlered, \circlepeach, \circlebeige\}) if it better predicts the observed text (\textit{conditional perplexity}) and the explanation is more internally consistent (\textit{posterior predictive}). Our paper is divided into two parts: in the first part (i.e. Section \ref{subsct:source_schemata} and Section \ref{sct:classifier-training}), we introduce the different schemata we will compare -- i.e. the top half of this diagram. In the first part (i.e. Section  \ref{sct:comparing_schemata} and Section \ref{sct:predicting_schemata}) we determine the right schema for a datum among competing schemata  -- i.e. the bottom half of this diagram -- and, given minimal information about a document, we show that we can predict what schema \textit{should} be used.
				% In the figure above, two \textit{plans} are different sequences of shapes, either squares and circles. We adapt 8 schemata to describe different plans and develop two metrics to favor one plan over another: \textit{conditional perplexity} \cite{airoldi2016improving} helps us measure how well a plan corresponds with the observed text and \textit{posterior predictive} \cite{spangher2023identifying} helps us measure how internally coherent a plan is.
			}
			\label{fig:cover-photo}
		\end{figure*}
		%
		%
		% A \textit{plan} is a sequence of actions made by a writer: 
		%  Inspired by classical methods to compare Bayesian topic models, we develop metrics to test document \textit{plans} in autoregressive settings. 
		%
		As language models (LMs) become more proficient at long-form text generation and incorporate resources \cite{lewis2020retrieval} and tools \cite{schick2023toolformer} to support their writing, recent work has shown that planning before writing is essential \cite{lecun2022path, spangher2023sequentially, park2023longstory}. However, supervised datasets to support learning and studying plans are few: they are difficult or expensive to collect, synthetic, or narrowly tailored to specific domains \cite{zhou2023webarena}.
		
		One approach to collecting diverse planning data is to observe natural scenarios in which planning has already occurred. In this work, we consider one such real-world scenario: source selection by human journalists. %Journalists use a variety of informational sources to inform storytelling. 
		Consider the article shown in Table \ref{tbl:intro-example-article}. The author shares her plan\footnote{Plan: \url{https://nyti.ms/3Tay92f} [paraphrased]. Final article: \url{https://nyti.ms/486I11u}, see Table \ref{tbl:intro-example-article}.}:
		
		\begin{quote}
			\textit{NJ schools are teaching climate change in elementary school. We wanted to understand: how are \textbf{teachers} educating children? How do \textbf{parents} and \textbf{kids} feel? Is there \textbf{pushback}?}
		\end{quote}
		
		As can be seen, the journalist planned, before writing, the different kinds of sources (e.g. teachers, kids) she wished to use. \textit{Why did she choose these groups?} Was it: A. to include varied social groups? % (i.e. \textit{stance}-based plan)? 
		B. to capture different sides of an issue?% (i.e. \textit{affiliation}-based plan)?
		%
		% planning Source selection, as illustrated above, is one of many planning decisions humans make, yet, attempts to 
		% have been critiqued for not planning ahead \cite{lecun2022path} or producing structured output \cite{spangher2023sequentially,  chakrabarty2023art} and even attempts to 
		% instill planning in LMs \cite{} fall short: creative tasks often lack a well-defined objective, so we have no good way of comparing plans, and plans made by humans are mostly not observable in the final text. \textbf{(2) Much work exists examining sourcing patterns in journalism \cite{winter2014question, hertzum2022journalists}, however no quantitative measure exists to understand the decisions journalists make on an article-by-article basis.} Sourcing decisions are important for representation and agenda-setting \cite{manninen2017sourcing}, and provide an important case-study of human planning. % Tools exist to aid in source discovery \cite{wang2021journalistic}, and our.
		
		Answering this question, we argue, allows us to infer why she chose each source. If the answer is A, we can infer, then, that the writer probably chose her sources because each fell into a different social group. If the answer is B, the sources were more likely chosen because each agreed or disagreed with the main event. Table \ref{tbl:intro-example-article} shows this duality. Establishing $P(A) > P(B)$ means we can better infer why each source was used, allowing us to collect plans from natural text data.
		
		Now, the core problem in this endeavor emerges: a document's plan is not typically observable. We directly address this and show that \textit{we can differentiate between plans in naturally observed text}. %We start by establishing two, simple metrics for comparing document-level plans. 
		Inspired by latent variable modeling approaches \cite{airoldi2016improving}, we uncover a document's most likely plan on the following basis: a proposed plan better describes a document's actual plan if it gives more information about the completed document. We introduce simple metrics for this goal: conditional perplexity and posterior predictive likelihood, in Figure \ref{fig:cover-photo} (Section \ref{subsect:schema_criticism}).
		
		Next, to create a straightforward setting to demonstrate the power of these metrics, we work with professional journalists from multiple major news organizations to identify planning approaches they regularly take. We operationalize these as schemata, or explanatory frameworks under which each source in the news article is assigned to a different discrete label (e.g. in the \textit{affiliation} schema, for example, the source-categories would be \textit{Government}, \textit{Media}...). We adapt five schemata from parallel tasks and introduce three novel schemata to better describe sourcing criterion. We implement our schemata by annotating over 600 news articles with 4,922 sources and training supervised classifiers. We validate our approach with these journalists: \textbf{they deem the plans we infer as correct with $>.74$ F1 score.}
		
		%These schemata capture broad aspects of how information relates both \textit{within} a document (e.g. stance detection \cite{hardalov2021cross}) as well \textit{extrinsically}: (e.g. retrieval \cite{spangher2023identifying}). % We annotate 4,922 sourcees across 600 articles and build classifiers for these schema, showing that we can model them with reasonable accuracy.
		%
		% Having curated these different approaches, we seek to compare them against each other. By viewing a document's source-categorization under different schemata as different latent-variable assignments,  %This frames schema-selection as Bayesian model-selection, building off recent work unifying latent variable modeling with large language modeling \cite{deng2022model, wang2023large}, \cite{wei2022chain}.
		%
		% Finally, we conduct a corpus-wide analysis and find that a source's \textit{social affiliation} and \textit{stance} optimally explain plans in most documents. However, for certain kinds of documents, e.g. factually dense topics like ``Science'', textual entailment (NLI) \cite{dagan2005pascal} provides a useful  structure. 
		
		Finally, the choice of schema, we find, can be predicted with moderate accuracy using only the headline of the article (ROC=.67), opening the door to new computational journalism tooling. % Finally, \textit{are these 8 schemata enough?} We extensively baseline against multiple latent variable models, which we build, and show that we cannot beat these schemata.
		
		In sum, our contributions are threefold:
		
		\begin{itemize}
			\item We frame \textit{source-type planning} as a lens through which to study planning in writing.
			\item We collect 8 different plan descriptions, or \textit{schemata} (5 existing and 3 we develop \textbf{with professional journalists}). We build a pipeline to extract sources from 4 million news articles and categorize them, building a large public dataset called \textit{NewsSources}.
			\item We introduce two novel metrics: \textit{conditional perplexity} and \textit{posterior predictive} to compare plans. We find that different plans are optimal for different topics. Further, we show that the right plan can be predicted with .67 ROC given just the headline.%, opening the door to advances in generative planning.
		\end{itemize}
		
		With this work, we hope to inspire further unsupervised inferences in document generation. Studying journalistic decision-making is important for understanding our information ecosystem \cite{winter2014question,
			manninen2017sourcing, debutts2024reporting}, can lead to important computational journalism tools \cite{quinonez2024new} and presents a real-world case-study in planning.

		\section{Source Categorization}
		\label{sct:problem_statement}
		\subsection{Problem Statement}
		\label{subsect:problem_statement}
		
		Our central question is: why did the writer select sources $s_1, s_2, s_3...$ for document $d$? Intuitively, let's say we read an article on a controversial topic. Let's suppose we observe that it contains many opposing viewpoints: some sources in the article ``agree'' with the main topic and others ``disagree''. We can conclude that the writer probably chose sources on the basis of their \textit{stance} \cite{hardalov2021cross} (or their opinion-based support)  rather than another explanation, like their \textit{discourse} role (which describes their narrative function). 
		
		More abstractly, we describe source-selection as a generative process: first, journalists plan \textit{how} they will choose sources (i.e. the \textit{set} of $k$ categories sources will fall into), then they choose sources, each falling into 1-of-$k$ categories. Different plans, or categorizations, are possible (e.g. see Figure \ref{tbl:intro-example-article}): \textit{the ``right'' plan is the one that best predicts the final document.}
		% Stated as a generative pseudo-story \cite{koller2009probabilistic}: 
		% \begin{quote}
			% \textit{``For a story topic $H$, sample $M$ sources, each with type $z \sim p(Z)$. For each source, sample words $x \sim p(w | z, H)$.% Find $\hat{Z} = \arg \max_{z \in Z} p(w | z, H)$
				% ''}. 
			% \end{quote}
		
\begin{figure*}[t]
			\centering
			% Answer: [trim={left bottom right top},clip]
			% \includegraphics[trim={4cm 0 4cm 0}, clip=True, angle=270, width=.6\linewidth]{figures/schema-collection-diagram.pdf}
			\includegraphics[width=.9\linewidth]{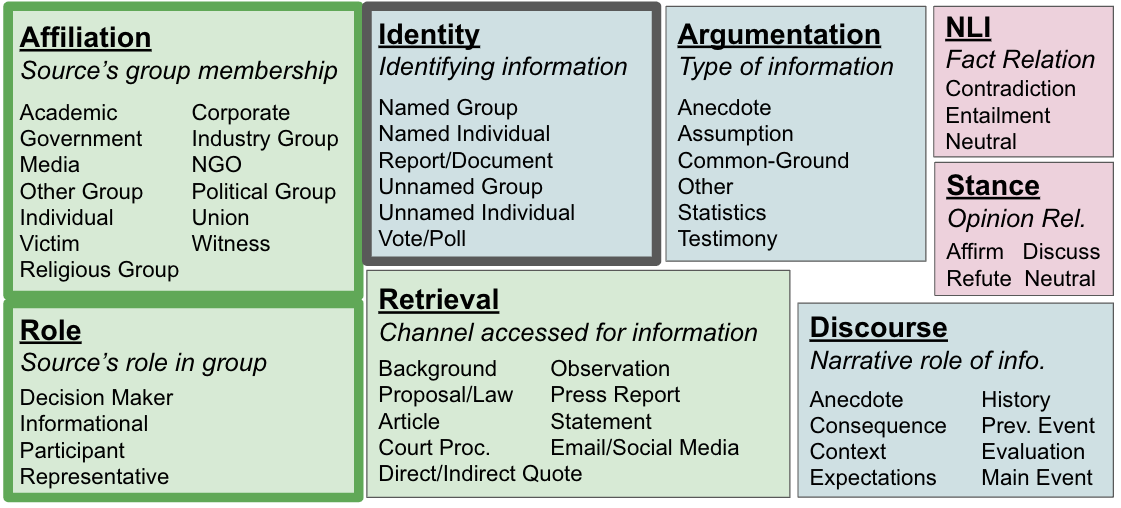}
			\caption{Label-sets for source-planning schemata. 
				%%%%%%%%
				\hlgreen{\textbf{Extrinsic Source Schemata}} Affiliation, Role and Retrieval-method \cite{spangher2023identifying} capture characteristics of sources \textit{extrinsic} to their usage in the document. 
				%%%%%%%%
				\hlazure{\textbf{Functional Source Schemata:}} Argumentation \cite{al2016news}, Discourse \cite{choubey2020discourse} and Identity capture functional narrative role of sources. 
				%%%%%%%%
				\hlpurple{\textbf{Debate-Oriented Schemata}}: Natural Language Inference (NLI) \cite{dagan2005pascal} and Stance \cite{hardalov2021cross} capture the role of sources in encompassing multiple sides. The three novel schemata we introduce are shown with borders: Affiliation, Identity and Role. For definitions, see Appendix \ref{app:schema_definitions}.
			}
			\label{fig:source-schemata}
		\end{figure*}

		Each plan, or categorizations, is specified by a \textit{schema}. For the 8 schema used in this work, see Figure \ref{fig:source-schemata}. To apply a schema to a document, we frame an approach consisting of two components: (1) an attribution function, $a$:
		\begin{align}
			a(s) = q \in Q_d \text{ for } s \in d
			% \\
			% q = \{s_{1,d}^{(q)}, ... s_{n,d}^{(q)}\}
			\label{eq:attribution-function}
		\end{align}
		\noindent introduced in \newcite{spangher2023identifying}, which maps each sentence $s$ in document $d$ to a source $Q_d = \{q_1^{(d)}, ... q_k^{(d)}\}$\footnote{These sources are referenced in $d$. There is no consideration of document-independent sources.} and (2) a classifier, $c$: 
		\begin{align}
			c_Z(s_1^{(q)}, ... s_n^{(q)}) = z \in Z
		\end{align}
		which takes as input a sequence of sentences attributed to source $q^{(d)}$ %(and optionally $h$, a headline or summary of the article) 
		and assigns a type $z \in Z$ for schema $Z$. %Taken together, $c_Z$ and $a$ give us a learned estimate of the posterior $p(z | x)$.
		
		This supervised framing is not typical in latent-variable settings; the choice of $z$ and the \textit{meaning} of $Z$ are typically jointly learned without supervision. However, learned latent spaces often do not correspond well to theoretical schemata \cite{chang2009reading}, and supervision has been shown to be helpful with planning \cite{wei2022chain}. On the other hand, supervised models trained on different schemata are challenging to compare, especially when different architectures are optimal for each schema. A latent-variable framework here is ideal: comparing different graphical models \cite{bamman2013learning, bamman2014unsupervised} \textit{necessitates} comparing different schemata, as each run of a latent variable model produces a different schema.% Thus methods \cite{wallach2009evaluation, airoldi2016improving}, to evaluate \textit{which latent variable assignment describes the observed data the best}, give us an apples-to-apples approach for determining \textit{which schema is better}. % here we wish to compare \textit{schemata}. %, or categories of possible assignments. 
		
		\subsection{Comparing \textit{Plans}, or Schemata}
		\label{subsect:schema_criticism}
		% We note that ``schema criticism'' is parallel to
		We can compare plans in two ways: (1) how well do they explain each observed document? and (2) how structurally consistent are they?
		
		% While \textit{conditional perplexity} $p(x | z)$ tells us how well a given $z$ explains $x$, we also wish to understand . reasoning about the likelihood of a \textit{sequence} of latent variables, $\hat{z}$, tells us how well $z$ as we are assuming them to be fixed. 
		
		\paragraph{Explainability} A primary criterion for a \textit{plan} is for it to explain the observed data well. To measure this, we use \textit{conditional perplexity}\footnote{We abuse notation here, using $p$ as both probability and perplexity: $p(x) = \exp \{-\mathbb{E} \log p (x_i | x_{<i}) \}$.}
		\begin{equation}
			p(x | z)
			\label{eq:conditional_perplexity}
		\end{equation}
		which measures the uncertainty of observed data, $x$, given a latent structure, $z$. Measuring $p(x|z)$ for different $z$ (fixing $x$) allows us to compare $z$. Conditional perplexity is a novel metric we introduce, inspired by metrics to evaluate latent unsupervised models, like the ``left-to-right'' algorithm introduced by \cite{airoldi2016improving}.  \footnote{We note that the term, \textit{conditional perplexity}, was originally introduced by \newcite{zhou1998word} to compare machine-translation pairs. In their case, both $x$ and $z$ are observable; as such, they do not evaluate latent structures, and their usage is not comparable to ours.}
		
		% -- here, we assume access to learned classifiers for $p(z | x)$. % we can fix $z$ and learn a model to directly calculate $p(x | z)$.  
		% We describe in Section \ref{} how we implement posterior predictive and conditional perplexity for schema criticism. 
		
		\paragraph{Structural Likelihood:} A second basic criterion for a latent structure to be useful is for it be consistent, which is a predicate for learnability. We assess the consistency of a set of assignments, $z$, by calculating the \textit{posterior predictive}: 
		\begin{equation}
			p(z | z_{-}, x)
			\label{eq:posterior_predictive}
		\end{equation}
		
		\noindent \newcite{deng2022model} exploring using full joint distribution, $p(z)$, \textit{latent perplexity}, to evaluate the structure text $x$ produced by generative language models (``\textit{model criticism}''). We simplify using the full distribution and instead evaluate the conditional predictive to study document structure. This, we find in early experiments, is easier to learn and thus helps us differentiate different $Z$ better (``\textit{schema criticism}'').\footnote{Our work is inspired by \newcite{spangher2023identifying}'s work, where $z$ was the choice of specific source, rather than a general source-type. However, they had no concept of a ``schema'' to group sources.
		} Now, we describe our schemata. %  In our case, we are primarily in measuring the choice of $Z$, or performing ``\textit{schema criticism}'' 
		% Also, we find, we do not need to calculate the entire joint distribution $p(z)$. 
		
		For an illustration of each metric, please refer to Figure \ref{fig:cover-photo}. The overall goal of the metrics is to determine \textit{which schema, or labeling of sources, best explains the observed news article}. As the figure shows, if schema A describes an article better than schema B, then labels assigned to each source under schema A (e.g. in Figure \ref{fig:cover-photo}: squares, \squareblue, \squarepurple, \squareteel) will outperform labels assigned under Schema B (e.g. circles, \circlebeige, \circlepeach, \circlered).
		
		\subsection{Source Schemata}
		\label{subsct:source_schemata}
		
		Our schemata, or plans, are shown in Figure \ref{fig:source-schemata}. We collect 8 schemata to compare, including three we introduce: \textit{\ul{Identity, Affiliation and Role}}.  Each schema provides a set of labels, which each describe sources used in a news article. Again, our hypothesis is that the schema which \textit{best predicts the observed text of the article} is the one the journalist most likely adhered to while planning the article (Section \ref{sct:comparing_schemata}). 
		% Two of the 8 schemata are  \textbf{debate-oriented} schemata (i.e. they describe how sources relate to the main topic of the article), three are \textbf{functional} (i.e. they describe the role sources serve in the overall flow of the story), and three can be considered \textbf{extrinsic} schemata (i.e. they describe how sources fit into societal structures). 
		See Appendix \ref{app:schema_definitions} for more details and definitions for each schema. 
		
		We note that \textbf{\textit{none}} of these schemata are complete and that real-world plans likely have elements outside of any one schema (or are combinations of multiple schema). However, this demonstration is important, we argue, to prove that we \textit{can} differentiate between purely latent plans in long-form text. We now introduce each schema:
		
		\paragraph{Debate-Oriented Schemata}
		Both the \textit{Stance} and \textit{NLI} schemata are debate-oriented schemata. They each capture the relation between the information a source provides and the main idea of the article. %: a \textit{premise} (\textbf{p}) and a \textit{hypothesis} (\textbf{h}). 
		\textit{NLI} \cite{dagan2005pascal} captures factual relations between text, while \textit{Stance} \cite{hardalov2021cross} captures opinion-based relations
		% \footnote{\newcite{reddy2021newsclaims} views these as the same.}
		. A text pair may be factually consistent and thus be classified as ``Entailment'' under a \textit{NLI} schema, but express different opinions and be classified as ``Refute'' under \textit{Stance}. In our setting, we relate article's headline with the source's attributable information. 
		These schemata say a writer uses sources for the purpose of expanding or rebutting information in the narrative, offering different perspectives and broadening the main idea. 
		
		\paragraph{Functional Source Schemata}
		
		The following schemata: \textit{Argumentation}, \textit{Discourse} and \textit{Identity} all capture the role a source plays in the overall narrative construction of the article. For instance, a source might provide a ``Statistic'' for a well-formed argument (\textit{Argumentation} \cite{al2016news}), or ``Background'' for a reader to help contextualize (\textit{Discourse} \cite{choubey2020discourse}). \textit{Identity}, a novel schema, captures how the reader identifies the source. For example, a ``Named Individual'' is identifiable to a reader, whereas an ``Unnamed Individual'' is not. As identified in  \newcite{sullivan2016tightening} and our journalist collaborators, this can be a strategic planning choice: some articles are about sensitive topics and need unnamed sources. % (see Appendix \ref{app:cases}).
		%because the information provided is vital to the story.% See Appendix \ref{} for more information.
		
		\begin{table}[]
			% \small 
			\centering
			\begin{tabularx}{\linewidth}{XrXr}
				\toprule
				Schema & Macro-F1 & Schema & Macro-F1\\
				\cmidrule(lr){1-2}\cmidrule(lr){3-4}
				Argumentation & 68.3 & Retrieval & 61.3 \\
				NLI & 55.2 & Identity & 67.2 \\
				Stance & 57.1 & Affiliation & 53.3 \\
				Discourse & 56.1 & Role & 58.1 \\
				\bottomrule
			\end{tabularx}
			\caption{Classification f1 score, macro-averaged, for the 8 schemata. We achieve moderate classification scores for each of schema. In Section 2, when we compare schemata, we account for classification acc. differences by introducing noise to higher-performing classifiers.}
			\label{tab:cls_accuracy}
		\end{table}
		
		\paragraph{Extrinsic Source Schemata}
		
		\textit{Affiliation}, \textit{Role} and \textit{Retrieval} schemata serve to characterize attributes of sources external to the news article. They either capture aspect about how sources exist as entities in society (\textit{Affiliation}, \textit{Role}), or the informational channel through which is was retrieved (\textit{Retrieval}). 
		Stories often implicate social groups \cite{mclean2019narrative}, such as ``academia" or ``government.'' Those group identities are extrinsic to the story's architecture but important for the selection of sources. Sources may be selected because they represent a group (i.e. \textit{Affiliation}) or because their group position is important within the story's narrative (e.g. ``participants'' in the events, i.e. \textit{Role}). \textit{Retrieval}, introduced by \newcite{spangher2023identifying}, captures the channel through which the information was found. Although these schemata are news-focused, we challenge the reader to imagine ones that might exist in other fields. For instance, a machine learning article might compare models selected via, say, a \textit{Community} schema: each from \textit{open-source}, \textit{academic} and \textit{industry research} communities.% They each represent different threads in society which the article seeks to unify.%\footnote{We anticipate that these schemata lend themselves well to retrieval via different existing knowledge bases \cite{}}.
		
		\section{Building a Silver-Standard Dataset of Different Possible Plans}
		
		The schemata described in the previous section give us theoretical frameworks for identifying writers' plans. To \textit{compare} schemata and select the schema that best describes a document, we must first create a dataset where informational sources are labeled \textit{according to each schema}. We describe that process in this section.
		
		\subsection{Dataset Construction and Annotation}
		\label{sct:dataset_construction}
		
		We obtain the NewsEdits dataset \cite{spangher2022newsedits}, which consists of 4 million news articles, and extract sources using a methodology developed by \newcite{spangher2023identifying}, which authors established was state-of-the-art for this task. This dataset spans 12 different news sources (e.g. BBC, NYTimes, etc.) over a period of 15 years (2006-2021). For our experiments, we sample $90,000$ news articles that are long and contain more than $3$ sources (on average, the articles contain $\sim 7.5$ sources). Then, we annotate to collect training data and build classifiers to categorize these sources. We described those processes now. 
		
		% We find that 47\% of sentences in our documents can be attributed to sources, and documents each contain an average of 7.5 +-/5 sources. These statistics are comparable to those reported by \newcite{spangher2023identifying}. %\footnote{Published between 2021-2023}
		
		% We annotate sources under each of our new schemata. 
		We recruited two annotators, one an undergraduate and the other a former journalist. The former journalist trained the undergraduate for 1 month to identify and label sources, then, they independently labeled 425 sources in 50 articles with each schema to calculate agreement, scoring $\kappa = .63, .76, .84$ on \textit{Affiliation}, \textit{Role} and \textit{Identity} labels. They then labeled 4,922 sources in 600 articles with each schema, labeling roughly equal amounts. Finally, they jointly labeled $100$ sources in $25$ documents with the other schemata for evaluation data over 1 month, with $\kappa \geq .54$, \textit{all in the range of moderate to substantial agreement} \cite{landis1977measurement}.

		\begin{table*}
			% \small 
			\centering
			\begin{tabular}{lr rrr rrr}
				\toprule
				& & \multicolumn{3}{c}{Conditional Perplexity $p(x | z)$} & \multicolumn{3}{c}{Posterior Predictive $p(\hat{z} | z_{-}, x)$}\\
				Schema        & $n$ & PPL & $\Delta$ base-k ($\downarrow$) & $\Delta$ base-r ($\downarrow$) & F1 & $\div$ base-k ($\uparrow$) & $\div$ base-r ($\uparrow$) \\
				\cmidrule(lr){1-2} \cmidrule(lr){3-5} \cmidrule(lr){6-8}
				NLI           & 3          & 22.8 & {\cellcolor[HTML]{F8C6C6}} \color[HTML]{000000} 0.62 & {\cellcolor[HTML]{F9FFF9}} \color[HTML]{000000}  -0.08 & 58.0 & {\cellcolor[HTML]{E9FEE9}} \color[HTML]{000000} 1.02** & {\cellcolor[HTML]{F5FFF5}} \color[HTML]{000000} 1.01 ** \\
				Stance        & 4          & 21.5 & {\cellcolor[HTML]{8AEF8A}} \color[HTML]{000000} -1.71 & {\cellcolor[HTML]{49BB49}} \color[HTML]{F1F1F1} -3.21** & 39.1 & {\cellcolor[HTML]{F17676}} \color[HTML]{F1F1F1} 0.88** & {\cellcolor[HTML]{F45C5C}} \color[HTML]{F1F1F1} 0.83 ** \\
				Role          & 4          & 22.3 & {\cellcolor[HTML]{F9FFF9}} \color[HTML]{000000} -0.06 & {\cellcolor[HTML]{E5FEE5}} \color[HTML]{000000} -0.33** & 38.7 & {\cellcolor[HTML]{92F692}} \color[HTML]{000000} 1.11** & {\cellcolor[HTML]{98FB98}} \color[HTML]{000000} 1.10 ** \\
				Identity      & 6          & 21.8 & {\cellcolor[HTML]{E1FEE1}} \color[HTML]{000000} -0.42 & {\cellcolor[HTML]{B8FCB8}} \color[HTML]{000000} -0.94 & 25.0 & {\cellcolor[HTML]{FDFFFD}} \color[HTML]{000000} 1.00 & {\cellcolor[HTML]{7EE67E}} \color[HTML]{000000} 1.15 ** \\
				Argumentation & 6          & 21.7 & {\cellcolor[HTML]{D9FED9}} \color[HTML]{000000} -0.52 & {\cellcolor[HTML]{B0FCB0}} \color[HTML]{000000} -1.04 & 30.7 & {\cellcolor[HTML]{98FB98}} \color[HTML]{000000} 1.10 ** & {\cellcolor[HTML]{8EF38E}} \color[HTML]{000000} 1.12 ** \\
				Discourse     & 8          & 22.3 & {\cellcolor[HTML]{F9D0D0}} \color[HTML]{000000} 0.54 & {\cellcolor[HTML]{C8FDC8}} \color[HTML]{000000}  -0.75 & 19.2 & {\cellcolor[HTML]{C0FDC0}} \color[HTML]{000000} 1.06 ** & {\cellcolor[HTML]{ACFCAC}} \color[HTML]{000000} 1.08 ** \\
				Retrieval     & 10         & 23.7 & {\cellcolor[HTML]{F07E7E}} \color[HTML]{F1F1F1} 1.47 & {\cellcolor[HTML]{FBDFDF}} \color[HTML]{000000}   0.36 & 15.8 & {\cellcolor[HTML]{98FB98}} \color[HTML]{000000} 1.10 ** & {\cellcolor[HTML]{8EF38E}} \color[HTML]{000000} 1.12 ** \\
				Affiliation   & 14         & 20.5 & {\cellcolor[HTML]{79E279}} \color[HTML]{000000} -2.11** & {\cellcolor[HTML]{50C150}} \color[HTML]{F1F1F1} -3.04** & 10.5 & {\cellcolor[HTML]{47B947}} \color[HTML]{F1F1F1} 1.26 ** & {\cellcolor[HTML]{7EE67E}} \color[HTML]{000000} 1.16 ** \\
				% \midrule
				% latent var. model & 14 & 22.06 & {\cellcolor[HTML]{D5FDD5}} \color[HTML]{000000} -0.58 & {\cellcolor[HTML]{93F793}} \color[HTML]{000000} -1.51 \\
				% topic-model-words-joined & 22.07 & 14 & 22.64 & 23.57 \\
				% topic-model-words-separated & 22.24 & 14 & 22.64 & 23.57 \\
				\bottomrule
			\end{tabular}
			\caption{Comparing our schemata against each other. In the first set of experiments, we show \textit{conditional perplexity} results, which tell us how well each schema explains the document text. Shown is PPL (the mean perplexity per schema), $\Delta kmeans$ (PPL - avg. perplexity of kmeans) and $\Delta random$ (PPL - avg. perplexity of the random trial). Statistical significance ($p<.05$) via a $t$-test calculated over perplexity values is shown via **. Higher perplexities mean worse predictive power, so the more negative the $\Delta$, the better.
				In the second set of experiments, we show \textit{posterior predictive} results, measured via micro F1-score. We show F1 (f1-score per schema), $\div$ kmeans (F1 / f1-score of kmeans), $\div$ random (F1 / f1-score of random trial). Statistical significance ($p<.05$) via a $t$-test calculated over 500-sample bootstrapped f1-scores is shown via **.}
			\label{tbl:results}
		\end{table*}
		
		\subsection{Training Classifiers to Label Sources}
		\label{sct:classifier-training}
		
		We train classifiers to label sources under each schema. Unless specified, we use a sequence classifier using RoBERTa-base with self-attention pooling, as in \newcite{spangher2021multitask}. We deliberately chose smaller models to scale to large amounts of articles. We will open-source all of the classifiers trained in this paper.
		
		\paragraph{\textit{Affiliation}, \textit{Role}, \textit{Identity}} We use our annotations to train classifiers which take as input all sentences attributable to source $q$ and output a label in each schema, or $p(t | s_1^{(q)} \oplus ... \oplus s_n^{(q)})$.
		
		\paragraph{\textit{Argumentation}, \textit{Retrieval}, \textit{Discourse}} We use datasets, without modification, that were directly released by the authors. Each is labeled on a sentence-level, on news and opinion datasets. We train classifiers to label each sentence of the news article, $s$. Then, for each source $q$, we assign a single label, $y$, with the most mutual information\footnote{$\arg\max_y p(y | q) / p(y)$)} across sentences attributed to that source, $s_1^{(q)}, ... s_n^{(q)}$.
		
		\paragraph{\textit{NLI, Stance}} We use an NLI classifier trained by \newcite{williams-etal-2020-anlizing} to label each sentence attributed to source $q$ as a separate hypothesis%, as in \newcite{laban2022summac}
		, and the article's headline as the premise. We use mutual information to assign a single label.
		
		We create a stance training dataset by aggregating several news-focused stance datasets\footnote{FNC-1 \cite{pomerleau2017fake}, Perspectrum \cite{chen2019seeing}, ARC \cite{habernal2017argument}, Emergent \cite{ferreira2016emergent} and NewsClaims \cite{reddy2021newsclaims}. We filter these sets to include premises and hypothesis $\geq$ 10 words and $\leq$ 2 sentences.}. %Data aggregation for stance detection inspired by: \cite{hardalov2021cross, schiller2021stance}}. 
	We then fine-tune GPT3.5-turbo\footnote{We use OpenAI's GPT3.5-turbo fine-tuning endpoint, as of November 16, 2023.} to label news data and label 60,000 news articles. We distill a $T5$ model with this data (Table \ref{tab:cls_accuracy} shows T5's performance).
	
	\subsection{Classification Results}
	\label{sct:cls_eda}
	As shown in Table \ref{tab:cls_accuracy}, we model schemata within a range of f1-scores $\in (53.3, 67.2)$, showing moderate success in learning each schema\footnote{When using these classifier outputs for evaluating plans, in the next section, we introduce noise (i.e. random label-swapping), so that all have the same accuracy.}. 
	These scores are middle-range and likely not useful on their own; we would certainly have achieved higher scores with more state-of-the-art methods. However, we note \textit{these classifiers are being used for comparative, explanatory purposes, so their efficacy lies in how well they help us compare plans}, as we will explore in the next section. 
	
	\section{Comparing Schemata}
	\label{sct:comparing_schemata}
	
	We are now ready to explore how well these schemata explain source selection in documents. We start by describing our experiments, then baselines, and finally results. All experiments in this section are based on the $90,000$ news articles filtered from NewsEdits, labeled as described in the previous section. We split $80,000$/$10,000$ train/eval.
	
	\subsection{Implementing Planning Metrics}
	\label{subsct:experiments}
	We now describe how we implement the metrics introduced in Section \ref{subsect:schema_criticism}: (1) \textit{conditional perplexity} and (2) \textit{posterior predictive}. 
	% , Equation \ref{eq:conditional_perplexity})
	% , Equation \ref{eq:posterior_predictive})
	
	% 
	
	\paragraph{Conditional Perplexity} To measure \textit{conditional perplexity}, $p(x | z)$%(Equation \ref{eq:conditional_perplexity})
	, we fine-tune GPT2-base models \cite{radford2019language} to take in it's prompt a sequence of latent variables, each for a different source, and \textit{then assess likelihood of the observed article text}.\footnote{We note that this formulation has overlaps with recent work seeking to learn latent plans \cite{deng2022model, wang2023large, wei2022chain}.} This is similar to measuring  \textit{vanilla perplexity} on observed text, except: (1) we provide latent variables as conditioning (2) by fixing the model used and varying the labels, \textit{we are measuring the signal given by each set of different labels}. 
	Our template for GPT2 is: 
	\begin{quote}
		\texttt{
			\colorbox{red!20}{
				$\langle$h$\rangle$ $h$ $\langle$l$\rangle$ (1) $l_1$ (2) $l_2$...$\langle$t$\rangle$} 
			\colorbox{green!20}{(1) $s_{1}^{(q_1)} ... s_{n}^{(q_1)}$ (2)... 
			}
		}
	\end{quote}
	
	\colorbox{red!20}{Red} is the prompt, or conditioning, and \colorbox{green!20}{green} is the text over which we calculate perplexity. \texttt{<tokens>}  (e.g. ``(1)'', ``$\langle$text$\rangle$'') are structural markers while variables $l, h, s$ are article-specific. $h$ is the headline, $l_i$ is the label for source $i$ and $s_{1}^{(q_1)} ... s_{n}^{(q_1)}$ are the sentences attributable to source $i$. \textit{\ul{We do not use GPT2 for generation, but for comparative purposes, to compare the likelihood of observed article text under each schema.}} We note that this implements Eq. \ref{eq:conditional_perplexity} only if we assuming \colorbox{green!20}{green} preserves the meaning of $x$, the article text. Our data processing (Section \ref{sct:dataset_construction}), based on high-accuracy source-extraction models \cite{spangher2023identifying}, gives us confidence in this.\footnote{Initial experiments show that text markers are essential for the model to learn structural cues. However, they also provide their own signal (e.g. on the number of sources). To reduce the effects of these artifacts, we use a technique called \textit{negative prompting} \cite{sanchez2023stay}. Specifically, we calculate perplexity on the \textit{altered} logits, $P_{\gamma} = \gamma \log p(x | z) - (1 - \gamma) \log p(x | \hat{z})$, where $\hat{z}$ is a shuffled version of the latent variables. Since textual markers remain the same in the prompt for $z$ and $\hat{z}$, this removes markers' predictive power.
	}
	
	\paragraph{Posterior Predictive} To learn the \textit{posterior predictive} (Equation \ref{eq:posterior_predictive}), we train a BERT-based classification model \cite{devlin2018bert} to take the article's headline and a sequence of source-types \textit{with a one randomly held out}. We then seek to predict \textit{that} source-type, and evaluate using F1-score. Additionally, we follow \newcite{spangher2023identifying}'s observation that some sources are \textit{more important} (i.e. have more information attributed). We model the posterior predictive among the 4 sources per article with the most sentences attributed to them.
	
	\subsection{Baselines}
	
	Vanilla perplexity does not always provide accurate model comparisons \cite{meister2021language, oh2022comparison} because it can be affected by irrelevant factors, like tokenization scheme. We hypothesized that the dimensionality of each schema's latent space might also have an effect \cite{lu2017vocabulary}; larger latent spaces tend to assign lower probabilities to each point. Thus, we benchmark each schema against baselines with similar latent dimensions.
	
	\paragraph{Base-r, or Random baseline}. We generate $k$ unique identifiers\footnote{Using MD5 hashes, from python's \texttt{uuid} library.}, and randomly assign one to each source in each document. $k$ is set to match the number of labels in the schema being compared to.
	
	\paragraph{Base-k, or Kmeans baseline}. We first embed sources as paragraph-embeddings using Sentence BERT \cite{reimers2019sentence} \footnote{Specifically, \texttt{microsoft/mpnet-base}'s model \url{https://www.sbert.net/docs/pretrained_models.html}.} Then, we cluster all sources across documents into $k$ clusters using the kmeans algorithm \cite{likas2003global}, where $k$ is set to match the number of labels in the schema being compared to. We assign each source it's cluster number.% it was assigned. 
	
	% \paragraph{Latent Variable Model}  We hypothesize that kmeans may be a poor unsupervised baseline, as cluster assignment might be confounded by topical aspects of the documents rather than the functional role of the sources. We adapt a Bayesian hierarchical model introduced by \cite{spangher2021don} designed to separate topical and functional components in text.
	% . We design a hierarchical latent model with the following components: a document-type per document, a source-type per source, and a ``word-switch'', based on \newcite{peng2014learning}, that separates (1) document-specific, topical words from (2) source-specific, functional words. 
	% We specify the model and variations we tested in Appendix \ref{app:topic-model}, including the Gibbs-sampler samplers derived. Because of the slow run-time, we do not run multiple trials.%, however we hope in future work to expand this \cite{bingham2019pyro}.
	
	\subsection{Results and Discussion}
	\label{sec:ppl_results}
	
	As shown in Table \ref{tbl:results}, the supervised schemata mostly have have lower conditional perplexity than their random and unsupervised kmeans baselines. However, only the \textit{Stance}, \textit{Affiliation} and \textit{Role} schemata improve significantly (at $p<.001$), and the \textit{Role} schema's performance increase is minor. \textit{Retrieval} has a statistically significant less explainability relative to it's baselines. 
	
	There is a simple reason for why some schemata have either the same or more conditional perplexity compared to their baselines: they lack explainability over the text of the document, but are not random and thus might lead to overfitting. We examine examples and find that \textit{Retrieval} does not impact wording as expected: writers make efforts to convey information similarly whether it was obtained via a quote, document or a statement.
	
	We face a dilemma: in generating these schemata, we chose \textit{Retrieval} because we assumed it was an important planning criterion. However, our results indicate that it holds little explanatory power. \textit{Is it possible that some plans do not get reflected in the text of the document?} 
	
	To address this question, we assign $\hat{Z} = \arg \min_{Z} p(x | z)$, the schema for each datapoint with the lowest perplexity, using scores calculated in the prior section\footnote{across the dataset used for validation, or 5,000 articles}, we calculate the lowest-perplexity schema. Table \ref{tbl:argmaxppl-distribution} shows the distribution of such articles. We then task 2 expert journalists with assigning their \textit{own} guess about which schema best describes the planning for the particular article, for 120 articles. \textbf{We observe an F1-score of 74, indicating a high degree of agreement.}
	
	% There are two reasons for this: (1) a small number of examples are very high perplexity, and this shifts the distribution significantly (when considering median statistics, as shown in Appendix \ref{app:eda}, the difference disappears.) (2) 
	
	Interestingly, we also observe statistically significant improvements of kmeans over random baselines in all cases (except $k=3$). In general, our baselines have lower variance in perplexity values than experimental schemata. This is not unexpected: as we will explore in the next section, we expect that some schemata will best explain only some articles, resulting in a greater range in performance. For more detailed comparisons, see Appendix \ref{app:eda}.
	
	Posterior predictive results generally show improvement across trials, with the \textit{Affiliation} trial showing the highest improvement over both baselines. This indicates that most tagsets are, to some degree, internally consistent and predictable. \textit{Stance} is the only exception, showing significantly lower f1 than even random baselines. This indicates that, although Stance is able to explain observed documents well (as observed by it's impact on conditional perplexity), it's not always predictable how it will applied. Perhaps this is indicative that writers do not know a-priori what sources will agree or disagree on any given topic before talking to them, and writers do not always actively seek out opposing sides.
	
	Finally, as another baseline, we implemented latent variable model. In initial experiments, it does not perform well. We show in Appendex \ref{app:topic-model} that the latent space learned by the model is sensible. Bayesian models are attractive for their ability to encode prior belief, and ideally they would make good baselines for a task like this, which interrogates latent structure. However, more work is needed to better align them to modern deep-learning baselines.
	
	\section{Predicting Schemata}
	\label{sct:predicting_schemata}
	
	Taken together, our observations from (1) Section \ref{sct:cls_eda}) indicate that schemata are largely unrelated and (2) Section \ref{sec:ppl_results} indicate that \textit{Stance} and \textit{Affiliation} both have similar explanatory power (although \textit{Stance} is less predictable). We next ask: which kinds of articles are better explained by one schema, and which are better explained by the other? If we can answer this question, we take steps towards being able to \textit{plan} source-selection via different schemata. Such a step could lead us towards better \textit{multi-document} retrieval techniques, by giving us axes to combine different documents into a retriever.
	
	In Table \ref{tbl:top-keywords}, we show topics that have low perplexity under the \textit{Stance} schema, compared with the \textit{Affiliation} schema (we calculate these by aggregating document-level perplexity across keywords assigned to each document in our dataset). As we can see, topics requiring greater degrees of debate, like ``Artificial Intelligence'', and ``Taylor Swift'' are favored under the \textit{Stance} Topic, while broader topics requiring many different social perspectives, like ``Culture'' and ``Freedom of Speech'' are favored under \textit{Affiliation}. We set up an experiment where we try to predict $\hat{Z} = \arg \min_{Z} p(x | z)$, the schema for each datapoint with the lowest perplexity. We downsample until assigned schemata, per articles, are balanced and train a simple linear classifier\footnote{Bag-of-words with logistic regression} to predict $\hat{Z}$. We get .67 ROC-AUC (or .23 f1-score). These results are tantalizing and offer the prospect of being able to \textit{better plan source retrieval} in computational journalism tools, by helping decide an axis on which to seek different sources. More work is needed to validate these results.
	
	\begin{table}[t]
		% \small 
		\begin{tabular}{ll}
			\toprule
			\textit{Stance} & \textit{Affiliation} \\
			\midrule
			Bush, George W & Freedom of Speech \\
			Swift, Taylor & 2020 Pres. Election \\
			Data-Mining & Jazz \\
			Artificial Intelligence & Ships and Shipping \\
			Rumors/Misinfo. & United States Military \\
			Illegal Immigration & Culture (Arts) \\
			Social Media & Mississippi \\
			\bottomrule
		\end{tabular}
		\caption{Top keywords associated with articles favored by stance or affiliation. Keywords are manually assigned by news editors}
		\label{tbl:top-keywords}
	\end{table}
	
	% We build Gibbs Samplers to solve these model
	
	\section{Related Work}
	\label{scn:related-work}
	This work focuses on informational sources in news articles and is part of a broader field of character-based analysis in text.
	
	\subsection{Latent Variable Persona Modeling} Our work is inspired by earlier work in persona-type latent variable modeling \cite{bamman2013learning, card-etal-2016-analyzing, spangher2021don}. Authors model characters in text as mixtures of topics. We both seek to learn and reason about about latent character-types, but their line of work takes an unsupervised approach. We show that supervised schemata outperform unsupervised.
	
	\subsection{Multi-Document Retrieval} 
	
	In multiple settings -- e.g. multi-document QA \cite{pereira2023visconde}, multi-document summarization \cite{shapira2021extending}, retrieval-augmented generation \cite{lewis2020retrieval} -- information \textit{from a single source} is assumed to be insufficient to meet a user's needs. In typical information retrieval settings, the goal is to retrieve a single document closest to the query \cite{page1998pagerank}. Despite earlier work in multi-document retrieval \cite{zhai2015beyond, yu2023search}, in settings where \textit{multiple sources are needed}, on the other hand, retrieval goals are not clearly understood\footnote{As \newcite{pereira2023visconde} states, \textit{``retrievers are the main bottleneck''} for well-performing multi-document systems.}. Our work attempts to clarify this, and can be seen as a step towards better retrieval planning.
	
	\begin{table}[t]
		% \small 
		\centering
		\begin{tabularx}{\linewidth}{XXXX}
			\toprule
			Affiliation & 41.7\% & Argument. & 1.2\% \\
			Identity & 22.7\% & Discourse & 1.1\% \\
			Stance & 17.7\% & NLI & 1.1\% \\
			Role & 13.4\% & Retrieval & 1.1\% \\
			\bottomrule
		\end{tabularx}
		\caption{Proportion of our validation dataset favored by one schema, i.e. $\hat{Z} = \arg \max_{Z} p(x | z)$}
		\label{tbl:argmaxppl-distribution}
	\end{table}
	
	\subsection{Planning in Language Models} Along the line of the previous point, chain-of-thought reasoning  \cite{wei2022chain} and in context learning (ICL), summarized in \cite{sanchez2023stay}, can be seen as latent-variable processes. Indeed, work in this vein is exploring latent-variable modeling for ICL example selection \cite{wang2024bayesian}. Our work, in particular the \textit{conditional perplexity} formulation and it's implementation, can be seen as a way of comparing different chain-of-thought plans as they relate to document planning.
	
	\subsection{Computational Journalism} Computational journalism seeks to apply computational techniques to assist journalists in reporting. Researchers have sought to improve detection of incongruent information \cite{chesney-etal-2017-incongruent}, detect misinformation \cite{pisarevskaya-2017-deception} and false claims made in news articles \cite{adair2017progress}. Such work can improve readers' trust in news. Our work takes steps towards understanding plans, or schemata, in news articles. As such, further work in this direction might yield deeper, more latent critiques for identifying untrustworthy articles.
	
	Another vein in computational journalism aims at improving journalists' story-writing abilities. One direction analyses news article revision logs \cite{tamori-etal-2017-analyzing} as a step towards automatic revision systems. Other research in this area seeks to identify and recommend relevant angles that have not been written yet for a trending story \cite{cucchiarelli-etal-2017-write}. Yet another direction aims to improve headline-writing by suggesting catchy headlines \cite{szymanski2017helping}. We see our source-modeling as relevant in this direction: mixture modeling of sources in documents can possibly identify gaps in stories and assess which sources to include. 
	
	Within this broad field, our work aims at aiding journalists by leading towards machine-in-the-loop systems. Overview, for instance, is a tool that helps investigative journalists comb through large corpora \cite{brehmer2014overview}. Workbench is another tool by the same authors aiming to facilitate web scraping and data exploration \cite{workbench}. Work by \newcite{diakopoulos2010diamonds} aims to surface social media posts that are \textit{unique} and \textit{relevant}. Our work is especially relevant in this vein. We envision characterizations of source types being combined with knowledge graphs to lead to similar tools for finding relevant sources, and suggesting sources to add to a story.
	
	\section{Conclusions}
	
	In conclusion, we explore ways of thinking about sourcing in human writing. We compare 8 schemata of source categorization, and adapt novel ways of comparing them. We find, overall, that \textit{affiliation} and \textit{stance} schemata help explain sourcing the best, and we can predict which is most useful with moderate accuracy. Our work lays the ground work for a larger discussion of discovering plans made by humans in naturally generated documents. It also takes us steps towards tools that might be useful to journalists. Naturally, our work is a simplification of the real human processes guiding source selection; these categories are non-exclusive and inexhaustive. We hope by framing these problems we can spur further research in this area.
	
	\section{Limitations}
	
	A central limitation to our work is that the datasets we used to train our models are all in English. As mentioned previously, we used English language sources from \citet{spangher2022newsedits}'s \textit{NewsEdits} dataset, which consists of sources such as nytimes.com, bbc.com, washingtonpost.com, etc. Thus, we must view our work with the important caveat that non-Western news outlets may not follow the same source-usage patterns and discourse structures in writing their news articles as outlets from other regions. We might face extraction and labeling biases if we were to attempt to do such work in other languages.
	
	Another limitation of our work is that we only considered 8 supervised schemata. While we worked closely with journalists to develop these schemata and attempted to make them as comprehensive and useful as possible, it's entirely possible, in fact probable, that these 8 schemata do not describe sources that well. As mentioned in the main body, we fully anticipate that more work needs to be done to determine further, more optimal schemata. And it's likely,  ultimately, that unsupervised approaches to developing more nuanced plans are desirable. 
	
	Furthermore, the metrics we evaluated are approximate and depend on schemata learned by ML models. Both of these facts could incentivize biased models. However, we attempted to mitigate this by conducting an experiment afterwards with journalists to blindly label articles.
	
	Our annotation approach was done only two annotators, in a master-apprentice style and hence might be skewed in distribution. However, because the master was an experienced journalist with many years of newsroom experience at a major newsroom, we took their tagging to be gold-standard.
	
	\section{Ethics Statement}
	
	\subsection{Risks}
	
	Since we constructed our datasets on well-trusted news outlets, we assumed that every informational sentence was factual, to the best of the journalist's ability, and honestly constructed. We have no guarantees that our classification systems would work in a setting where a journalist was acting adversarially.
	
	There is a risk that, if planning works and natural language generation works advance, it could fuel actors that wish to use it to plan misinformation and propaganda. Any step towards making generated news article more human-like risks us being less able to detect and stop them. Misinformation is not new to our media ecosystem, \cite{boyd2018characterizing, spangher2020characterizing}. We have not experimented how our classifiers would function in such a domain. There is work using discourse-structure to identify misinformation \cite{abbas2022politicizing}, and this could be useful in a source-attribution pipeline to mitigate such risks.
	
	We used OpenAI Finetuning to train the GPT3 variants. We recognize that OpenAI is not transparent about its training process, and this might reduce the reproducibility of our process. We also recognize that OpenAI owns the models we fine-tuned, and thus we cannot release them publicly. Both of these thrusts are anti-science and anti-openness and we disagree with them on principle. We tried where possible to train open-sourced versions, as mentioned in the text. 
	
	\subsection{Licensing}
	
	The dataset we used, \textit{NewsEdits} \cite{spangher2022newsedits}, is released academically. Authors claim that they received permission from the publishers to release their dataset, and it was published as a dataset resource in NAACL 2023. We have had lawyers at a major media company ascertain that this dataset was low risk for copyright infringement.
	
	\subsection{Computational Resources}
	
	The experiments in our paper required computational resources. We used 64 12GB NVIDIA 2080 GPUs. We designed all our models to run on 1 GPU, so they did not need to utilize model or data-parallelism. However, we still need to recognize that not all researchers have access to this type of equipment.
	
	We used Huggingface models for our predictive tasks, and will release the code of all the custom architectures that we constructed. Our models do not exceed 300 million parameters.
	
	\subsection{Annotators}
	
	We recruited annotators from our educational institutions. They consented to the experiment in exchange for mentoring and acknowledgement in the final paper. One is an undergraduate student, and the other is a former journalist. Both annotators are male. Both identify as cis-gender. The annotation conducted for this work was deemed exempt from review by our Institutional Review Board.

	\bibliography{main}
	\bibliographystyle{acl_natbib}
	
	\newpage
	\appendix

	\addcontentsline{toc}{section}{Appendix} % Add the appendix text to the document TOC
	\part{Appendix} % Start the appendix part
%	\parttoc % Insert the appendix TOC

	In Appendix \ref{app:methodological_details}, we include more, precise detail about our experimental methods. Then, Appendix \ref{app:eda}, we present more exploratory analysis to support our experiments, including comparisons between schemata. In Appendix \ref{app:schema_definitions}, we give a more complete set of definitions for the labels in each schema. In Appendix \ref{app:topic-model}, we define the unsupervised latent variable models we use as baselines, including providing details on their implementation. %Finally, in Appendix \ref{app:cases}, we show how the schemata we introduce can be generally useful for corpora analysis by demonstrating two case-studies: observational differences following editorial policy changes, and political changes following the 2016 presidential election.
	
	% Before we start, here is another example of a news article along with the description, by the journalist, of the sources categories they started to investigate.
	
	\section{Additional Methodological Details}
	\label{app:methodological_details}
	
	\subsection{Source Extraction}
	\label{subsct:attribution_method}
	Before classifying sources, we first need to learn an attribution function (Equation \ref{eq:attribution-function}) to identify the set of sources in news articles. \newcite{spangher2023identifying} introduced a large source attribution dataset, but their models are either closed (i.e. GPT-based) or underperforming. So, we train a high-performing open-source model using their dataset. We fine-tune GPT3.5-turbo \footnote{As of November 30th, 2023.}, achieving a prediction accuracy of 74.5\% on their test data\footnote{Lower than the reported 83.0\% accuracy of their Curie model. We formulate a different, batched prompt aimed at retrieving more data.}. Then, we label a large silver-standard dataset of 30,000 news articles and distill a BERT-base span-labeling model, described in \cite{vaucher2021quotebank}, with an accuracy of 74.0\%.\footnote{All models will be released.} We use this model to score a large corpus of $90,000$ news articles from the NewsEdits corpus \cite{spangher2022newsedits}. We find that 47\% of sentences in our documents can be attributed to sources, and documents each contain an average of 7.5 +-/5 sources. These statistics are comparable to those reported by \newcite{spangher2023identifying}. %\footnote{Published between 2021-2023}

	\section{Exploratory Data Analysis}
	\label{app:eda}
	
	We explore more nuances of our schemata, including comparative analyses. We start by showing a view of $\hat{Z}$, the conditions under which a schema best explains the observed results. In Tables \ref{tbl:top-keywords-1} and \ref{tbl:top-keywords-2}, we show an extension of Table \ref{tbl:top-keywords} in the main body: we show favored keywords across all schemata. (Note that in contrast to Table \ref{tbl:top-keywords}, we restrict the keywords we consider to a tighter range). When topics require a mixture of different information types, like statistics, testimony, etc. \textit{Argumentation} is favored. When story-telling is on topics like ``Travel'', ``Education'', ``Quarantine (Life and Culture)'', where it incorporates background, history, analysis, expectation, \textit{Discourse} is favored. In Table \ref{tbl:top_sections}, we show the top \textit{Affiliations} per section of the newspaper, based on the NYT LDC corpus \cite{sandhaus2008new}. 
	
	\begin{figure}[t]
		\centering
		\includegraphics[width=\linewidth]{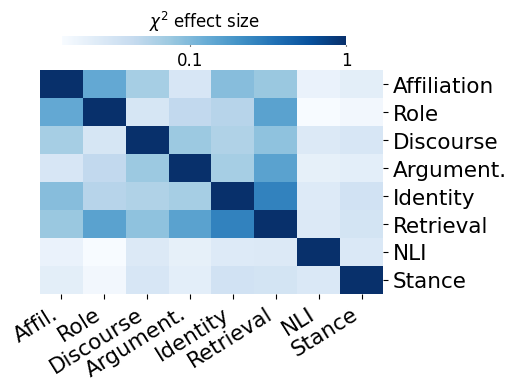}
		\caption{Correlation between 8 schemata, measured as Cramer's V \cite{cramer1999mathematical}, or the effect-size measurement of the $\chi^2$ test of independence.}
		\label{fig:schema-correlation}
	\end{figure}
	
	\begin{figure}
		\centering
		\includegraphics[width=\linewidth]{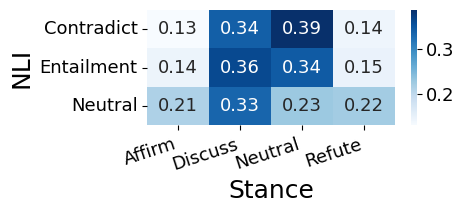}
		\caption{Stance and NLI schemata definitions are not very aligned. We show conditional probability of labels in each schema, $p(x|y)$ where $x=$ Stance and $y=$ NLI.}
		\label{fig:stance_vs_nli}
	\end{figure}
	
	\begin{table*}[t]
		\begin{tabularx}{\linewidth}{>{\hangindent=1em}X>{\hangindent=1em}X>{\hangindent=1em}X>{\hangindent=1em}X}
			\toprule
			Affiliation & Argumentation & Discourse & NLI \\
			\midrule
			Inflation (Economics) & Race and Ethnicity & Travel and Vacations & Deaths (Fatalities) \\
			\arrayrulecolor{lightgray} \cmidrule[.01pt](lr){1-1}\cmidrule(lr){2-2}\cmidrule(lr){3-3}\cmidrule(lr){4-4}
			Writing and Writers & Books and Literature & Quarantine (Life and Culture) & Murders, Homicides \\
			\cmidrule(lr){1-1}\cmidrule(lr){2-2}\cmidrule(lr){3-3}\cmidrule(lr){4-4}
			United States Economy & Demonstrations, Protests and Riots & Education (K-12) & Law and Legislation \\
			\cmidrule(lr){1-1}\cmidrule(lr){2-2}\cmidrule(lr){3-3}\cmidrule(lr){4-4}
			Race and Ethnicity & Travel and Vacations & Fashion and Apparel & States (US) \\
			\cmidrule(lr){1-1}\cmidrule(lr){2-2}\cmidrule(lr){3-3}\cmidrule(lr){4-4}
			Disease Rates & Suits and Litigation & Murders, Homicides & Science \\
			\cmidrule(lr){1-1}\cmidrule(lr){2-2}\cmidrule(lr){3-3}\cmidrule(lr){4-4}
			Real Estate and Housing (Residential) & Senate & Great Britain & Politics and Government \\
			\cmidrule(lr){1-1}\cmidrule(lr){2-2}\cmidrule(lr){3-3}\cmidrule(lr){4-4}
			China & United States International Relations & Deaths (Fatalities) & Personal Profile \\
			\cmidrule(lr){1-1}\cmidrule(lr){2-2}\cmidrule(lr){3-3}\cmidrule(lr){4-4}
			Supreme Court (US) & Deaths (Fatalities) & Pop and Rock Music & Children/ Childhood \\
			\cmidrule(lr){1-1}\cmidrule(lr){2-2}\cmidrule(lr){3-3}\cmidrule(lr){4-4}
			Ukraine & Labor and Jobs & Demonstrations, Protests and Riots & China \\
			\arrayrulecolor{black}
			\bottomrule
		\end{tabularx}
		\caption{Keyword topics that are best explained (i.e. have the lowest conditional perplexity) by the following schemata: Affiliation, Discourse, NLI. Broader topics, like ``Inflation'' which require sources from different backgrounds, favor Affiliation-based source selection, while topics integrating many different, possibly conflicting, facts, favor NLI-based selection.}
		\label{tbl:top-keywords-1}
	\end{table*}
	
	\begin{figure}
		\includegraphics[width=\linewidth]{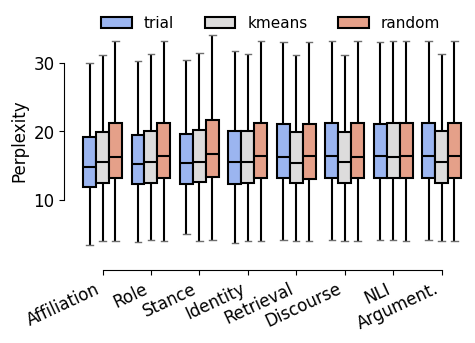}
		\caption{Distribution of conditional perplexity measurements across different experimental groups.}
		\label{fig:median-ppls-across-groups}
	\end{figure}
	
	\begin{figure}
		\centering
		\subfloat[Relationship between the size of the labelset and perplexity for kmeans trials]{
			\includegraphics[width=\linewidth]{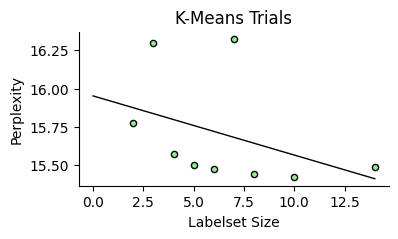}
			% \label{<figure1>}
		}\\
		\subfloat[Relational between the size of the labelset and perplexity for random trials.]{
			\includegraphics[width=\linewidth]{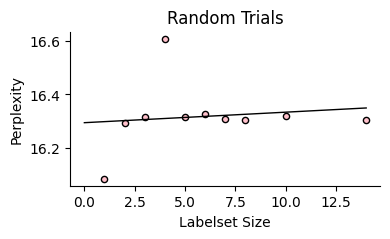}
			% \label{<figure1>}
		}\\
		\subfloat[Distribution over perplexity scores for all random trials and kmeans trials, compared.]{
			\includegraphics[width=\linewidth]{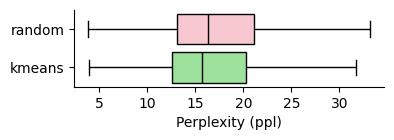}
			% \label{<figure1>}
		}    
		\caption{To explore the effects of labelset size, and confirm that conditional perplexity does align with basic intuitions, we compare Random trials and Kmeans trials across all of our labelset sizes.}
		\label{fig:kmeans-vs-random}
	\end{figure}
	
	\begin{figure}[t]
		\centering
		\includegraphics[width=\linewidth]{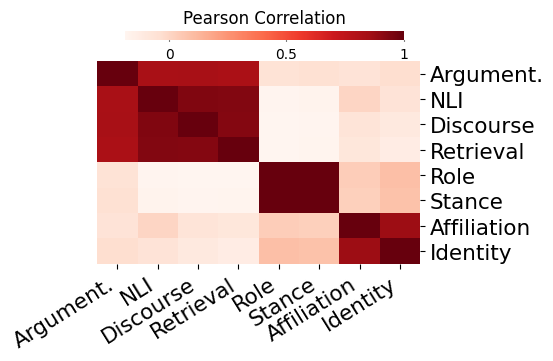}
		\caption{Pearson Correlation between conditional perplexity per document under different schemata.}
		\label{fig:ppl-correlation}
	\end{figure}
	
	\begin{table*}[t]
		\begin{tabularx}{\linewidth}{>{\hangindent=1em}X>{\hangindent=1em}X>{\hangindent=1em}X>{\hangindent=1em}X}
			\toprule
			Retrieval & Role & Identity & Stance \\
			\midrule
			Actors and Actresses & Inflation (Economics) & United States Economy & Midterm Elections (2022) \\
			\arrayrulecolor{lightgray} \cmidrule[.01pt](lr){1-1}\cmidrule(lr){2-2}\cmidrule(lr){3-3}\cmidrule(lr){4-4}
			Fashion and Apparel & House of Representatives & Disease Rates & Presidential Election of 2020 \\
			\cmidrule[.01pt](lr){1-1}\cmidrule(lr){2-2}\cmidrule(lr){3-3}\cmidrule(lr){4-4}
			Pop and Rock Music & Presidential Election of 2020 & Real Estate and Housing (Residential) & California \\
			\cmidrule[.01pt](lr){1-1}\cmidrule(lr){2-2}\cmidrule(lr){3-3}\cmidrule(lr){4-4}
			Elections & United States Economy & Movies & Storming of the US Capitol (Jan, 2021) \\
			\cmidrule[.01pt](lr){1-1}\cmidrule(lr){2-2}\cmidrule(lr){3-3}\cmidrule(lr){4-4}
			Personal Profile & Trump, Donald J & Education (K-12) & Vaccination and Immunization \\
			\cmidrule[.01pt](lr){1-1}\cmidrule(lr){2-2}\cmidrule(lr){3-3}\cmidrule(lr){4-4}
			Deaths (Fatalities) & Education (K-12) & Race and Ethnicity & News and News Media \\
			\cmidrule[.01pt](lr){1-1}\cmidrule(lr){2-2}\cmidrule(lr){3-3}\cmidrule(lr){4-4}
			Primaries and Caucuses & Elections, House of Representatives & Ukraine & United States Economy \\
			\cmidrule[.01pt](lr){1-1}\cmidrule(lr){2-2}\cmidrule(lr){3-3}\cmidrule(lr){4-4}
			Politics and Government & Supreme Court (US) & Trump, Donald J & Defense and Military Forces \\
			\cmidrule[.01pt](lr){1-1}\cmidrule(lr){2-2}\cmidrule(lr){3-3}\cmidrule(lr){4-4}
			Regulation and Deregulation of Industry & Computers and the Internet & Presidential Election of 2020 & Television \\
			\arrayrulecolor{black} \bottomrule
		\end{tabularx}
		\caption{Keyword topics that are best explained (i.e. have the lowest conditional perplexity) by the following schemata: Retrieval, Role, Identity, Stance. Political topics, like ``House of Representatives'' which often have a mixture of different roles, favor Role-based source selection, while polarizing topics like ``Storming of the US Capitol'' favor Stance.}
		\label{tbl:top-keywords-2}
	\end{table*}
	
	Next, we further explore the relation between different labelsets. In Figure \ref{fig:median-ppls-across-groups}, we show the same story as in Table \ref{tbl:results} in the Main Body, except with a broader view of the distributional shifts. As can be seen, by comparing differents between the means in Table \ref{tbl:results} and the medians in \ref{fig:median-ppls-across-groups}, we see that the effect of outliers is quite large, which reduces the significance we observe. In \ref{fig:ppl-correlation}, we show the correlation between perplexities across labelsets. We observe clusters in our schemata of particularly high correlation. Interestingly, this stands in contrast to Figure \ref{fig:schema-correlation}, which showed almost no relation between the tagsets. We suspect that outlier effects on perplexity (e.g. misspelled words, strange punctuation) has a high effect on relating different conditional perplexities, swamping the effects of the schema. This points to the caution in using perplexity as a metric; it must be well explored and appropriately baselined.
	
	In Figure \ref{fig:stance_vs_nli}, we explore more why NLI and Stance are not very related. It turns out that many of the factual categories can fall in any one of the opinion-based categories. A lot of ``Entailing'' facts under NLI, for example, might be the the basis of ``Discussion'' under Stance. This points to the need to be cautious when using NLI as a stand-in for Stance, as in \cite{reddy2021newsclaims}.
	
	In Figures \ref{fig:kmeans-vs-random}, we compare random and kmeans perplexities across the latent dimension size. Our experiments show that indeed, we are learning important cues about perplexity. As expected, ``Random'' assignments have almost no affect on the perplexity of the document, while ``kmeans'' assignments do. Increasing the dimensionality space of Kmeans, interestingly, \textit{decreases} the median perplexity, perhaps because the Kmeans algorithm is allowed to capture more and more meaningful semantic differences between sources.
	
	\begin{table}[t]
		\centering
		\begin{tabular}{lr rrr}
			\toprule
			Schema & n & H & \% Maj. & \% Min. \\
			\midrule
			Affiliation & 14 & 2.2 & {\cellcolor[HTML]{A6CEE4}} \color[HTML]{000000} 32.9 & {\cellcolor[HTML]{F6FAFF}} \color[HTML]{000000} 0.46 \\
			Role        &  4 & 1.0 & {\cellcolor[HTML]{56A0CE}} \color[HTML]{F1F1F1} 53.3 & {\cellcolor[HTML]{EEF5FC}} \color[HTML]{000000} 4.61 \\
			Identity    &  6 & 1.3 & {\cellcolor[HTML]{5AA2CF}} \color[HTML]{F1F1F1} 52.2 & {\cellcolor[HTML]{F6FAFF}} \color[HTML]{000000} 0.69 \\
			Argument.   &  6 & 1.1 & {\cellcolor[HTML]{3787C0}} \color[HTML]{F1F1F1} 62.9 & {\cellcolor[HTML]{F7FBFF}} \color[HTML]{000000} 0.22 \\
			NLI         &  3 & 1.1 & {\cellcolor[HTML]{89BEDC}} \color[HTML]{000000} 40.4 & {\cellcolor[HTML]{C8DCF0}} \color[HTML]{000000} 22.6 \\
			Stance      &  4 & 1.3 & {\cellcolor[HTML]{A0CBE2}} \color[HTML]{000000} 34.8 & {\cellcolor[HTML]{D7E6F5}} \color[HTML]{000000} 15.5 \\
			Discourse   &  8 & 1.9 & {\cellcolor[HTML]{B0D2E7}} \color[HTML]{000000} 30.0 & {\cellcolor[HTML]{F5FAFE}} \color[HTML]{000000} 1.09 \\
			Retrieval  &  10 & 2.0 & {\cellcolor[HTML]{CADEF0}} \color[HTML]{000000} 21.4 & {\cellcolor[HTML]{F7FBFF}} \color[HTML]{000000} 0.05 \\
			\bottomrule
		\end{tabular}
		\caption{Description of the size of each schema (n) and the class imbalance inherent in it, shown by: Entropy (H), \% Representation of the Majority class (\% Maj.) and \% Representation of the Minority class (\% Min.).}
		\label{tab:statistics}
	\end{table}

	Finally, we discuss label imbalances in our classification sets. We do not observe a strong correlation between the number of labels in a schema and the classification accuracy ($\rho=-.16$). As seen in Table \ref{tab:statistics}, many schemata are highly skewed, with, for example, the minority class in Argumentation (``common ground'') being present in less than .22\% of sources.
	Using our classifiers to label the news articles compiled in Section \ref{subsct:attribution_method}, we find that the schemata all offer different information. Figure \ref{fig:schema-correlation} shows the effect size of the $\chi^2$ independence test, a test ranging from $(0, 1)$ which measures the relatedness of two sets of categorical variables \cite{cramer1999mathematical}. The schemata are largely uncorrelated, with the highest correspondence being $\nu=.34$ between ``Identity'' and ``Retrieval''. We were surprised that NLI and Stance were not very related, as they have similar labelsets and have been used interchangeably \cite{reddy2021newsclaims}. This indicates that significant semantic differences exist between fact-relations and opinion-relations, resulting in different application of tags. We explore this in Appendix \ref{app:eda}.
	
	\begin{table*}[t]
		\begin{tabular}{llll}
			\toprule
			Newspaper Sections & \multicolumn{3}{c}{Proportion of Sources with each Label}  \\
			\midrule
			Arts & {\cellcolor[rgb]{0.298, 0.596, 0.773}} \color{white} Individual: 0.29 & {\cellcolor[rgb]{0.639, 0.737, 0.855}} \color{white} Media: 0.19 & {\cellcolor[rgb]{0.694, 0.761, 0.867}} \color{white} Witness: 0.17 \\
			Automobiles & {\cellcolor[rgb]{0.016, 0.388, 0.612}} \color{white} Corporate: 0.41 & {\cellcolor[rgb]{0.694, 0.761, 0.867}} \color{white} Witness: 0.17 & {\cellcolor[rgb]{0.839, 0.839, 0.910}} Media: 0.11 \\
			Books & {\cellcolor[rgb]{0.412, 0.643, 0.800}} \color{white} Individual: 0.26 & {\cellcolor[rgb]{0.639, 0.737, 0.855}} \color{white} Media: 0.19 & {\cellcolor[rgb]{0.667, 0.749, 0.863}} \color{white} Witness: 0.18 \\
			Business & {\cellcolor[rgb]{0.008, 0.220, 0.345}} \color{white} Corporate: 0.51 & {\cellcolor[rgb]{0.612, 0.725, 0.847}} \color{white} Government: 0.2 & {\cellcolor[rgb]{0.929, 0.906, 0.949}} Industry Group: 0.06 \\
			Dining and Wine & {\cellcolor[rgb]{0.333, 0.612, 0.780}} \color{white} Witness: 0.28 & {\cellcolor[rgb]{0.667, 0.749, 0.863}} \color{white} Individual: 0.18 & {\cellcolor[rgb]{0.694, 0.761, 0.867}} \color{white} Media: 0.17 \\
			Education & {\cellcolor[rgb]{0.063, 0.467, 0.702}} \color{white} Government: 0.36 & {\cellcolor[rgb]{0.639, 0.737, 0.855}} \color{white} Academic: 0.19 & {\cellcolor[rgb]{0.859, 0.851, 0.918}} Witness: 0.1 \\
			Front Page & {\cellcolor[rgb]{0.008, 0.220, 0.345}} \color{white} Government: 0.5 & {\cellcolor[rgb]{0.875, 0.867, 0.925}} Political Group: 0.09 & {\cellcolor[rgb]{0.894, 0.882, 0.933}} Corporate: 0.08 \\
			Health & {\cellcolor[rgb]{0.157, 0.529, 0.733}} \color{white} Government: 0.33 & {\cellcolor[rgb]{0.639, 0.737, 0.855}} \color{white} Academic: 0.19 & {\cellcolor[rgb]{0.824, 0.824, 0.906}} Corporate: 0.12 \\
			Home and Garden & {\cellcolor[rgb]{0.580, 0.710, 0.839}} \color{white} Individual: 0.21 & {\cellcolor[rgb]{0.639, 0.737, 0.855}} \color{white} Witness: 0.19 & {\cellcolor[rgb]{0.694, 0.761, 0.867}} \color{white} Corporate: 0.17 \\
			Job Market & {\cellcolor[rgb]{0.412, 0.643, 0.800}} \color{white} Corporate: 0.26 & {\cellcolor[rgb]{0.749, 0.788, 0.882}} Individual: 0.15 & {\cellcolor[rgb]{0.776, 0.800, 0.890}} Witness: 0.14 \\
			Magazine & {\cellcolor[rgb]{0.518, 0.686, 0.824}} \color{white} Witness: 0.23 & {\cellcolor[rgb]{0.612, 0.725, 0.847}} \color{white} Media: 0.2 & {\cellcolor[rgb]{0.667, 0.749, 0.863}} \color{white} Individual: 0.18 \\
			Movies & {\cellcolor[rgb]{0.333, 0.612, 0.780}} \color{white} Individual: 0.28 & {\cellcolor[rgb]{0.667, 0.749, 0.863}} \color{white} Media: 0.18 & {\cellcolor[rgb]{0.667, 0.749, 0.863}} \color{white} Witness: 0.18 \\
			New York and Region & {\cellcolor[rgb]{0.063, 0.467, 0.702}} \color{white} Government: 0.36 & {\cellcolor[rgb]{0.804, 0.812, 0.898}} Witness: 0.13 & {\cellcolor[rgb]{0.824, 0.824, 0.906}} Individual: 0.12 \\
			Obituaries & {\cellcolor[rgb]{0.667, 0.749, 0.863}} \color{white} Government: 0.18 & {\cellcolor[rgb]{0.667, 0.749, 0.863}} \color{white} Individual: 0.18 & {\cellcolor[rgb]{0.725, 0.776, 0.878}} \color{white} Media: 0.16 \\
			Opinion & {\cellcolor[rgb]{0.016, 0.361, 0.565}} \color{white} Government: 0.43 & {\cellcolor[rgb]{0.776, 0.800, 0.890}} Media: 0.14 & {\cellcolor[rgb]{0.824, 0.824, 0.906}} Witness: 0.12 \\
			Real Estate & {\cellcolor[rgb]{0.157, 0.529, 0.733}} \color{white} Corporate: 0.33 & {\cellcolor[rgb]{0.580, 0.710, 0.839}} \color{white} Government: 0.21 & {\cellcolor[rgb]{0.824, 0.824, 0.906}} Individual: 0.12 \\
			Science & {\cellcolor[rgb]{0.016, 0.404, 0.635}} \color{white} Academic: 0.4 & {\cellcolor[rgb]{0.639, 0.737, 0.855}} \color{white} Government: 0.19 & {\cellcolor[rgb]{0.859, 0.851, 0.918}} Corporate: 0.1 \\
			Sports & {\cellcolor[rgb]{0.016, 0.431, 0.675}} \color{white} Other Group: 0.38 & {\cellcolor[rgb]{0.749, 0.788, 0.882}} Individual: 0.15 & {\cellcolor[rgb]{0.776, 0.800, 0.890}} Witness: 0.14 \\
			Style & {\cellcolor[rgb]{0.518, 0.686, 0.824}} \color{white} Individual: 0.23 & {\cellcolor[rgb]{0.612, 0.725, 0.847}} \color{white} Witness: 0.2 & {\cellcolor[rgb]{0.694, 0.761, 0.867}} \color{white} Corporate: 0.17 \\
			Technology & {\cellcolor[rgb]{0.016, 0.388, 0.612}} \color{white} Corporate: 0.41 & {\cellcolor[rgb]{0.694, 0.761, 0.867}} \color{white} Government: 0.17 & {\cellcolor[rgb]{0.875, 0.867, 0.925}} Academic: 0.09 \\
			The Public Editor & {\cellcolor[rgb]{0.012, 0.345, 0.537}} \color{white} Media: 0.44 & {\cellcolor[rgb]{0.725, 0.776, 0.878}} \color{white} Individual: 0.16 & {\cellcolor[rgb]{0.725, 0.776, 0.878}} \color{white} Government: 0.16 \\
			Theater & {\cellcolor[rgb]{0.122, 0.506, 0.722}} \color{white} Individual: 0.34 & {\cellcolor[rgb]{0.667, 0.749, 0.863}} \color{white} Witness: 0.18 & {\cellcolor[rgb]{0.776, 0.800, 0.890}} Media: 0.14 \\
			Travel & {\cellcolor[rgb]{0.451, 0.659, 0.808}} \color{white} Witness: 0.25 & {\cellcolor[rgb]{0.580, 0.710, 0.839}} \color{white} Corporate: 0.21 & {\cellcolor[rgb]{0.749, 0.788, 0.882}} Government: 0.15 \\
			U.S. & {\cellcolor[rgb]{0.012, 0.345, 0.537}} \color{white} Government: 0.44 & {\cellcolor[rgb]{0.824, 0.824, 0.906}} Political Group: 0.12 & {\cellcolor[rgb]{0.894, 0.882, 0.933}} Academic: 0.08 \\
			Washington & {\cellcolor[rgb]{0.008, 0.220, 0.345}} \color{white} Government: 0.6 & {\cellcolor[rgb]{0.859, 0.851, 0.918}} Political Group: 0.1 & {\cellcolor[rgb]{0.894, 0.882, 0.933}} Media: 0.08 \\
			Week in Review & {\cellcolor[rgb]{0.031, 0.447, 0.694}} \color{white} Government: 0.37 & {\cellcolor[rgb]{0.839, 0.839, 0.910}} Academic: 0.11 & {\cellcolor[rgb]{0.859, 0.851, 0.918}} Media: 0.1 \\
			World & {\cellcolor[rgb]{0.008, 0.220, 0.345}} \color{white} Government: 0.54 & {\cellcolor[rgb]{0.875, 0.867, 0.925}} Media: 0.09 & {\cellcolor[rgb]{0.875, 0.867, 0.925}} Witness: 0.09 \\
			\bottomrule
		\end{tabular}
		\caption{Distribution over source-types with different \textit{Affiliation} tags, by newspaper section.}
		\label{tbl:top_sections}
	\end{table*}
	
	\section{Article Example}
	
	Here is an article example, annotated with different schemata definitions, along with a description by the journalist of why they pursued the sources they did.
	
	\begin{quote}
		\textit{We mined state and federal court paperwork. We went looking for [previous] stories. We called police and fire communications people to determine [events]. We found families for interviews about [the subjects'] lives.}\footnote{\url{https://www.nytimes.com/2017/01/23/insider/on-the-murder-beat-times-reporters-in-new-yorks-40th-precinct.html}}
	\end{quote}
	
	\begin{table}[t]
		\centering
		% \begin{tabular}{p{12.7cm}p{2.6cm}}		
			\begin{tabular}{p{7cm}} 
				\toprule
				\textbf{Headline: Services failed to prevent crime} \\
				\midrule
				\hspace {4mm} \_\_'s voice became a preoccupation of \_\_, who told the police that he heard her calling his name at night.  $\leftarrow$  \hlgreen{Government}, \hlpurple{Neutral} \\
				\hspace{4mm} ``Psychotic Disorder,'' detectives wrote in their report. \hspace{.15cm} $\leftarrow$ {\small \textit{labels:}} \hlgreen{Government}, \hlpurple{Refute} \\
				\hspace{4mm} ``She had a strong voice,'' said Carmen Martinez, 85, a neighbor. \hspace{.1cm} $\leftarrow$ {\small       \textit{}} \hlgreen{Witness}, \hlpurple{Neutral} \\
				\hspace{4mm}  Records show a string of government encounters failed to help \_\_ as his mental health deteriorated.  \hfill $\leftarrow$ {\small \textit{labels:}} \hlgreen{Government}, \hlpurple{Agree} \\
				\hspace{4mm}``This could have been able to be avoided,'' said \_\_'s lawyer.  \hfill $\leftarrow$ {\small \textit{labels:}} \hlgreen{Actor}, \hlpurple{Agree}  \\
				\bottomrule
			\end{tabular}
			\caption{Informational sources synthesized in a single news article\footnote{\url{https://www.nytimes.com/2016/02/19/nyregion/in-a-bronx-police-precinct-homicides-persist-as-crime-drops-elsewhere.html}}. Source categorizations under two different schemata: \hlgreen{affiliation} and \hlpurple{stance}. Our central question: \textit{\ul{which schema best characterizes the kinds of sources needed to tell this story?}}}
			\label{tbl:example-article}
		\end{table}
		
		\section{Further Schemata Definitions}
		\label{app:schema_definitions}
		
		Here we provide a deeper overview of each of the schemata that we used in our work, as well as definitions that we presented to the annotators during annotation.
		
		% A different category of sources do not belong to formal organizations. They are \underline{Individuals}: \textit{Actors}, \textit{Victims} and \textit{Witnesses}. These sources differ based on how active a role they take in the events around them: actors affect events around them, while witnesses and victims are neutral or affected by the events around them. Often, these sources cannot be directly reached and journalists seek proxies: family members, lawyers, doctors or spokespeople. 
		
		\begin{itemize}
			\item \textbf{Affiliation:} Which group the source belongs to.
			\begin{itemize}
				\item \textbf{Institutional:} The source belongs to a larger institution.
				\begin{enumerate}
					\item \textbf{Government:} Any source who executes the functions of or represents a government entity. \textit{(E.g. a politician, regulator, judge, political spokesman etc.)}
					\item \textbf{Corporate:} Any source who belongs to an organization in the private sector. \textit{(E.g. a corporate executive, worker, etc.)}
					\item \textbf{Non-Governmental Organization (NGO):} If the source belongs to a nonprofit organization that operates independently of a government. \textit{(E.g. a charity, think tank, non-academic research group.)}
					\item \textbf{Academic:} If the source belongs to an academic institution. Typically, these are professors or students and they serve an informational role, but they can be university administrators, provosts etc. if the story is specifically about academia.
					\item \textbf{Other Group:} If the source belongs or is acting on behalf of some group not captured by the above categories (please specify the group).
				\end{enumerate}
				\item \textbf{Individual:} The source does \textbf{NOT} belong to a larger institution.
				\begin{enumerate}
					\item \textbf{Actor:} If the source is an individual acting on their own. \textit{(E.g. an entrepreneur, main character, solo-acting terrorist.)}
					\item \textbf{Witness:} A source that is ancillary to events, but bears witness in either an active \textit{(e.g. protester, voter)} or inactive \textit{(i.e. bystander)} way.
					\item \textbf{Victim:} A source that is affected by events in the story, typically negatively.
					\item \textbf{Other:} Some other individual (please specify).
				\end{enumerate}
			\end{itemize}
			\item \textbf{Role:}
			\begin{enumerate}
				\item \textbf{Participant:} A source who is either directly making decisions on behalf of the entity they are affiliated with, or taking an active role somehow in the decision-making process.
				\item \textbf{Representative:} A source who is speaking on behalf of a \textit{Participant}.
				\item \textbf{Informational:} A source who is giving information on ongoing decisions or events in the world, but is not directly involved in them.
				\item \textbf{Other:} Some other role that we have not captured (please specify).
			\end{enumerate}
			\item \textbf{Role Status:}
			\begin{enumerate}
				\item \textbf{Current:} A source who is currently occupying the role and affiliation.
				\item \textbf{Former:} A source who \textit{used} to occupy the role and affiliation.
				\item \textbf{Other:} Some other status that we have not captured (please specify).
			\end{enumerate}
		\end{itemize}
		
		We note that \textbf{Rote Status} was a schema that we collected, but ultimately did not end up modeling.
		
		\section{Example GPT Prompts}
		\label{app:example_prompts}
		
		We give more examples for prompts. 
		
		\subsection{Source Attribution Prompts}
		
		In Section \ref{subsct:attribution_method}, we discuss training a GPT3.5-Turbo model with \newcite{spangher2023identifying}'s source attribution dataset to create more labeled datapoints, which we then distil into a BERT model. We train a batched model to save on costs. The prompt takes the following form:
		
		\texttt{
			\textbf{Input:}\\
			1. <sent 1>\\
			2. <sent 2>\\
			3. <sent 3>...\\
			\textbf{Response:}\\
			1. <attribution 1>\\
			2. <attribution 2>\\
			...
		}
		
		Here is an example:
		
		\texttt{
			\textbf{System Prompt:}\\
			You are a journalist's fact-checker who identifies sources providing information for each sentence. 
			The user will show you a sentences in an article and you'll respond with the source of the sentences. Consider the whole article and be sure to answer every question.\\
			Answer either by directly copying text in the article OR with "passive-voice" when a canonical source is clearly consulted OR "journalist" when a direct observation is made OR "No source" when no source is referenced, the information is vague, or the source is unclear.\\
			Do not make up names, or say anything that is not in the article besides those phrases above.\\
			\textbf{User Input:}
			1: BANGKOK — A plane carrying key senior Laotian government officials crashed Saturday morning, leaving at least four people dead, Laotian diplomats said Saturday.\\
			2: Killed in the crash were two top figures in the security apparatus of the authoritarian Lao  government: the deputy prime minister, Douangchay Phichit, and Thongbane Sengaphone, the minister of public security, according to two Lao diplomats.\\
			3: For a Communist party that relies on force and intimidation to stay in power, the loss of what were arguably the two most powerful people in the security apparatus was a significant blow.\\
			4: The governor of Vientiane province was also killed in the crash.\\
			5: In addition to his post as deputy prime minister Mr. Douangchay was defense minister and a member of the Politburo , the highest decision - making body of the Communist party.\\
			6: Mr. Thongbane, the public security head, was feared in the country and was said to be one of the officials leading a crackdown against dissent over the past year and half.\\
			\textbf{Response:}\\
			1: Laotian diplomats\\
			2: No Source\\
			3: No Source\\
			4: No Source\\
			5: No Source\\
			6: No Source\\
		}
		
		\subsection{Stance-Based Prompts}
		
		In Section \ref{sct:classifier-training} we discuss the prompts we formulated to do appropriate transfer learning from the stance datasets others have annotated to our news setting. Because in Stance detection, there are usually many claims made for each hypothesis, we used batched prompts to save costs, in the following form:
		
		\texttt{\textbf{Premise:} <premise>\\
			Claim: \\
			1. <claim 1>\\
			2. <claim 2>\\
			\textbf{Response:}
			1. <label 1>\\
			2. <label 2>\\
			...}
		
		Here is an example:
		
		\texttt{\textbf{System Prompt}:
			You are a journalist's assistant who spots opposing claims. The user will give you a premise and 5 claims. Respond to each one, in numbered order from 1 to 5, with a choice from: ['Neutral', 'Affirm', 'Discuss', 'Refute']. \\
			Don't say anything else, and be sure to answer each one.\\
			\textbf{User Prompt}\\
			Premise: 3-D printing will change the world. \\
			Claims:\\
			1: I can see 3D printing for prototypes, and some custom work. However manufacturing industries use thousands of plastics and thousands of metal alloys...\\
			2: Flash backwards to 1972, Colorado, where the newly enfranchised...\\
			3: This is precisely the way I feel about 3D printers...another way to fill the world with plastic junk that will end up in landfills, beaches, and yes, mountains and oceans. ...\\
			4: I am totally terrified with the thought of 3-D printed, non-traceable, guns and bullets in every thugs hands. May that never happen. But then Hiroshima did (bad thing)...\\
			5: Hate to point out an obvious solution is to tie the tax rate to unemployment....\\
			\textbf{Response:}\\
			1: Refute\\
			2: Neutral\\
			3: Refute\\
			4: Affirm\\
			5: Neutral}

		\subsection{GPT-2 Conditional Perplexity Prompts}
		
		In Section \ref{subsct:experiments}, we discuss crafting prompts for GPT2-base models in order to calculate conditional perplexity. We give the outline of our prompt. Here is an example:
		
		\texttt{
			Revelations from the artist’s autobiography threaten to cloud her new show at the San Francisco Museum of Modern Art. \\
			<labels>\\
			(1): NGO,\\
			(2): Media,\\
			(3): Media,\\
			(4): Media,\\
			(5): Corporate\\
			<text>\\
			(1): In a telephone interview on Tuesday, the museum\'s current director, Christopher Bedford , said he welcomed the opportunity to "be very outspoken about the museum\'s relationship to antiracism" and ... \\
			(2): Last week a Chronicle critic denounced the museum\'s decision to proceed with the show. \\
			(3): Its longest-serving curator, Gary Garrels, resigned in 2020 soon after a post quoted him saying, "Don\'t worry, we will definitely continue to collect white artists." \\
			(4): The website Hyperallergic surfaced those comments in June . \\
			(5): And its previous director, Neal Benezra, apologized to employees after removing critical comments from an Instagram post following the murder of George Floyd. \\
			(6): And the San Francisco Museum of Modern Art has been forced to reckon with what employees have called structural inequities around race. \\
			(7): The popular Japanese artist Yayoi Kusama, whose " Infinity Mirror Rooms " have brought lines around the block for one blockbuster exhibition after another, has...'}
		
		\section{Combining Different Schemata}
		
		We show how two schemata, \textit{Role} and \textit{Affiliation} may be naturally combined. One function of journalism is to interrogate the organizations powering our society. Thus, many sources are from \underline{Affiliations}: \textit{Government}, \textit{Corporations}, \textit{Universities}, \textit{Non-Governmental Organizations} (NGOs). And, they have different \textit{Roles} in these places. Journalists first seek to quote \textit{decision-makers} or \textit{participants}: presidents, CEOs, or senators. Sometimes decision-makers only comment though \textit{Representatives}: advisors, lawyers or spokespeople. These sources all typically provide knowledge of the inner-workings of an organization. Broader views are often sought from \textit{Informational} sources: experts in government or analysts in corporations; scholars in academia or researchers in NGOs. These sources usually provide broader perspectives on topics. Table \ref{tab:source-ontology} shows the intersection of these two schemata.
		
		\begin{table*}[t]
			\begin{tabular}{|l|l||l||l|l|l|}
				\cline{4-6}
				\multicolumn{3}{l|}{\multirow{2}{*}{}} & \multicolumn{3}{c|}{\textbf{Role}} \\
				\cline{4-6}
				\multicolumn{3}{l|}{} & \textit{Decision Maker} & \textit{Representative} & \textit{Informational}    \\
				\hhline{---|=|=|=|}
				\multirow{8}{*}{\rotatebox[origin=c]{90}{\textbf{Affiliation}}} & \multirow{5}{*}{\rotatebox[origin=c]{90}{\textit{Institutional}}} & \textit{Government} & President, Senator... & Appointee, Advisor... & Expert, Whistle-Blower... \\
				\cline{3-6}
				&& \textit{Corporate} & CEO, President... & Spokesman, Lawyer... & Analyst, Researcher... \\
				\cline{3-6}
				&& \textit{NGO}       & Director, Actor... & Spokesman, Lawyer... & Expert, Researcher... \\
				\cline{3-6}
				&& \textit{Academic}   & President, Actor... & Trustee, Lawyer...& Expert, Scientist... \\
				\cline{3-6}
				&& \textit{Group}      & Leader, Founder... & Member, Militia... & Casual, Bystander...\\
				\hhline{~|-||=||===|}
				& \multirow{3}{*}{\rotatebox[origin=c]{90}{\textit{Individ.}}}
				& \textit{Actor}      & Individual... & Doctor, Lawyer... & Family, Friends... \\
				\cline{3-6}
				&& \textit{Witness}   & Voter, Protestor... & Spokesman, Poll... & Bystander... \\
				\cline{3-6}
				&& \textit{Victim}    & Individual... & Lawyer, Advocate... & Family, Friends...\\
				\hline
			\end{tabular}
			\caption{Our source ontology: describes the affiliation and roles that each source can take. A \textit{source-type} is the concatenation of \textit{affiliation} and \textit{role}.}
			\label{tab:source-ontology}
		\end{table*}
		
		\section{Latent Variable Models}
		\label{app:topic-model}
		
		\begin{figure}[t]
			\centering
			\begin{tikzpicture}
				\tikzstyle{main}=[circle, minimum size = 6mm, thick, draw =black!80, node distance = 4mm]
				\tikzstyle{connect}=[-latex, thick]
				\tikzstyle{box}=[rectangle, draw=black!100]
				%
				% nodes
				%
				\node[main](Ht) [label=center:${H_T}$] { };
				\node[main](Pt) [below=of Ht, label=center:${P_T}$] { };
				\node[main](T) [right=of Pt, label=center:$T$] { };
				\node[main] (S) [above right=.8cm of T,label=center:$S$] { };
				\node[main] (Ps) [above=of S,label=center:$P_S$] { };
				\node[main] (Hs) [left=of Ps,label=center:$H_S$] { };
				\node[main] (z) [right=1.8cm of T,label=center:z] {};
				\node[main] (Pz) [below=.4cm of z,label=center:$P_z$] { };
				\node[main] (Hz) [right=of Pz,label=center:$H_z$] { };
				\node[main, fill = black!10] (w) [right=of z,label=center:w] { };
				\node[main] (Pw) [above=1.4cm of w,label=center:$P_w$] { };
				\node[main] (Hw) [left=of Pw,label=center:$H_w$] { };
				\node[main, fill = black!10] (gamma) [above=of z,label=center:$\gamma$] { };
				%
				% paths
				%
				\path
				(Ht) edge [connect] (Pt)
				(Pt) edge [connect] (T)
				(T) edge [connect] (S)
				(T) edge [connect] (z)
				(Ps) edge [connect] (S)
				(Hs) edge [connect] (Ps)
				(S) edge [connect] (z)
				(Pz) edge [connect] (z)
				(Hz) edge [connect] (Pz)
				(z) edge [connect] (w)
				(Pw) edge [connect] (w)
				(Hw) edge [connect] (Pw)
				(gamma) edge [connect] (z);
				%
				% plates
				% 
				% plate for N
				\node[rectangle, inner sep=-2.5mm, fit= (w) (gamma),label=below right:N, xshift=-.2mm] {};
				\node[rectangle, inner sep=2.5mm,draw=black!100, fit=(gamma) (w)] {};
				% plate for D
				\node[rectangle, inner sep=-1mm, fit= (T) (w) (gamma),label=below left:D, xshift=-1mm] {};
				\node[rectangle, inner sep=3.5mm,draw=black!100, fit=(T) (w) (gamma)] {};
				% plate for P
				\node[rectangle, inner sep=0mm, fit= (S),label=below right:S, xshift=-1mm] {};
				\node[rectangle, inner sep=3.0mm,draw=black!100, fit=(S)] {};
			\end{tikzpicture}
			\caption{Plate diagram for Source Topic Model}
			\label{fig:plateandstick}
		\end{figure}
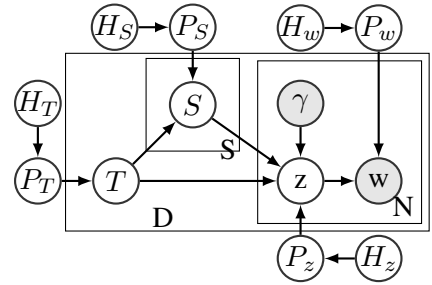
		
		As shown in Figure \ref{fig:plateandstick}, our model observes a switching variable, $\gamma$ and the words, $w$, in each document. The switching variable, $\gamma$ is inferred and takes one of two values: ``source word'' for words that are  associated with a source ``background'', for words that are not. 
		
		The model then infers source-type, $S$, document type $T$, and word-topic $z$. These variables are all categorical. All of the variables labeled $P_{.}$ in the diagram represent Dirichlet \textit{P}riors, while all of the variables labeled $H_{.}$ in the diagram represent Dirichlet \textit{H}yperpriors.
		
		Our generative story is as follows:
		
		For each document $d = 1,..., D$:
		
		\begin{enumerate}
			\itemsep-.2em
			\item Sample a document type $T_d \sim Cat(P_T)$
			\item For each source $s=1,...,S_{(d, n)}$ in document:
			\begin{enumerate}
				\item Sample source-type $S_s \sim Cat(P_S^{(T_d)})$
			\end{enumerate}
			\item For each word $w=1, ... N_w$ in document:
			\begin{enumerate}
				\item If $\gamma_{d, w} = $ ``source word'',		sample word-topic $z_{d, w} \sim Cat(P_z^{(S_s)})$
				\item If $\gamma_{d, w} = $ ``background'',
				sample word-topic $z_{d, w} \sim Cat(P_z^{(T_d)})$
				\item Sample word $w \sim Cat(z_{d, n})$
			\end{enumerate}
		\end{enumerate}
		
		% One can imagine the generative process our model specifies, informally, as expanding upon the scenario of the editor and the young reporter given in the introduction, with the following details added to the editor's request: ``\textit{there's been a car accident in Highland Park} [story type $T_d =$ `car accident'], \textit{go find me $3$ witnesses and $2$ police officers} [source-type $S_{d, 1}, S_{d, 2}, S_{d,3}$ = ``witness'', $S_{d, 4}, S_{d, 5} = $ ``police officer''], \textit{get me some quotes} [words $w_{d, w}$ where $\gamma_{d, w} = $ source-word] \textit{and write me a lead, a scene description and some background} [words $w_{d, w}$ where $\gamma_{d, w} = $ `background'].
		
		The key variables in our model, which we wish to infer, are the document type ($T_d$) for each document, and the source-type ($S_{(d, n)}$) for each source. 
		It is worth noting a key difference in our model architecture: \newcite{bamman2013learning} assume that there is an unbounded set of mixtures over person-types. In other words, in step 2, $S_s$ is drawn from a document-specific Dirichlet distribution, $P_S^{(d)}$. While followup work by \newcite{card-etal-2016-analyzing} extends \newcite{bamman2013learning}'s model to ameliorate this, \newcite{card-etal-2016-analyzing} do not place prior knowledge on the number of document types, and rather draw from a Chinese Restaurant Process.\footnote{\newcite{card-etal-2016-analyzing} do not make their code available for comparison.} We constraint the number of \textit{document-types}, anticipating in later work that we will bound news-article types into a set of common archetypes, much like we did for \textit{source-types}.
		
		Additionally, both previous models represent documents solely as mixtures of characters. Ours, on the other hand, allows the type of a news article, $T$, to be determined both by the mixture of sources present in that article, and the other words in that article. For example, a \textit{crime} article might have sources like a government official, a witness, and a victim's family member, but it might also include words like ``gun'', ``night'' and ``arrest'' that are not included in any of the source words.
		
		\subsection{Inference}
		
		We construct the joint probability and collapse out the Dirichlet variables: $P_w$, $P_z$, $P_S$, $P_T$ to solve a Gibbs sampler. 
		% We implement the following equations to sample features in our model:
		Next, we discuss the document-type, source-type, and word-topic inferences.
		
		\subsubsection{Document-Type inference}
		
		First, we sample a document-type $T_d \in {1, ..., T}$ for each document:
		
		\begin{equation}
			\begin{array}{c}
				p(T_d | T_{-d}, s, z, \gamma, H_T, H_S, H_Z) \propto \\
				(H_{TT_d} + c_{T_d, *}^{(-d)}) 
				\times 
				\prod_{s=1}^{S_d}
				\frac
				{(H_{Ss} + c_{T_d, s, *, *})}
				{(c_{T_d, *, *, *} + S H_{S})}
				\\ \times
				\prod_{j=1}^{N_T}
				\frac
				{(H_{zk} + c_{k, *, T_d, *})}
				{(c_{*,*,T_d,*} + K H_{z})}
			\end{array}
		\end{equation}
		
		\noindent where the first term in the product is the probability  attributed to document-type: $c_{T_d, *}^{(-d)}$ is the count of all documents with type $T_d$, not considering the current document $d$'s assignment. The second term is the probability attributed to source-type in a document: the product is over all sources in document $d$. Whereas $c_{T_d, s, *, *}$ is the count of all sources of type $s$ in documents of type $T_d$, and $c_{T_d, *, *, *}$ is the count of all sources of any time in documents of type $T_d$. The third term is the probability attributed to word-topics associated with the background word: the product is over all background words in document $d$. Here, $c_{k, *, T_d, *}$ is the count of all words with topic $k$ in document type $T_d$, and $c_{*, *, T_d, *}$ is the count of all words in documents of type $T_d$.
		
		\subsubsection{Source-Type Inference}
		
		Next, having assigned each document a type, $T_d$, we sample a source-type $S_{(d, n)} \in 1, ..., S$ for each source.
		
		\begin{equation}
			\begin{array}{c}
				p(S_{(d,n)} | S_{-(d, n)}, T, z, H_T, H_s, H_z) \propto \\
				(H_{SS_d} + c_{T_d, S_{(d,n)}, *, *}^{-(d,n)}) 
				\\ 
				\times
				\prod_{j=1}^{N_{S_{d, n}}}
				\frac
				{(H_{z} + c_{z_j, *, S_{(d, n)}, *, *})}
				{(c_{*, *, S_{(d, n)}, *, *} + K H_{z})}
			\end{array}
		\end{equation}
		
		The first term in the product is the probability attributed to the source-type: $c_{T_d, S_{(d, n)}, *, *}^{-(d, n)}$ is the count of all sources of type $S_{(d, n)}$ in documents of type $T_d$, not considering the current source's source-type assignment. The second term in the product is the probability attributed to word-topics of words assigned to the source: the product is over all words associated with source $n$ in document $d$. Here, $c_{z_j, *, S_{(d, n)}, *, *}$ is the count of all words with topic $z_j$ and source-type $S_{(d, n)}$, and $c_{*, *, S_{(d, n)}, *, *}$ is the count of all words associated with source-type $S_{(d,n)}$. 
		
		\subsubsection{Word-topic Inference}
		
		Finally, having assigned each document a document-type and source a source-type, we sample word-topics. For word $i, j$, if it is associated with sources ($\gamma_{i, j} = $ Source Word), we sample:
		
		\begin{equation}
			\begin{array}{c}
				p(z_{(i, j)} | z^{-(i, j)}, S, T, w, \gamma, H_w, H_S, H_T, H_z)
				\propto\\
				(c_{z_{i, j}, *, S_d, *, *}^{-(i, j)} + H_{zz_{i, j}}) 
				\times
				\frac
				{c_{z_{i, j}, *, w_{i, j}, *}^{-(i, j)} + H_{w}}
				{c_{z_{i, j}, *, *, *}^{-(i, j)} + V H_{w}}
			\end{array}
			\label{eqn:source-word-prob}
		\end{equation}
		
		The first term in the product is the word-topic probability:  $c_{z_{i, j}, *, S_d, *, *}^{-(i, j)}$ is the count of word-topics associated with source-type $S_d$, not considering the current word. The second term is the word probability: $c_{z_{i, j}, *, w_{i, j}, *}^{-(i, j)}$ is the count of words of type $w_{i, j}$ associated with word-topic $z_{i, j}$, and $c_{z_{i, j}, *, *, *}^{-(i, j)}$ is the count of all words associated with word-topic $z_{i, j}$.
		
		For word $i, j$, if it is associated with background word-topic ($\gamma_{i, j}$ = Background), we sample:
		
		\begin{equation}
			\begin{array}{c}
				p(z_{(i, j)} | z^{-(i, j)}, S, T, w, \gamma, H_w, H_S, H_T, H_z)
				\propto\\
				(c_{z_{i, j}, *, T_d, *}^{-(i, j)} + H_{zz_{i, j}}) 
				\times
				\frac
				{c_{z_{i, j}, *, w_{i, j}, *}^{-(i, j)} + H_{w} }
				{ c_{z_{i, j}, *, *, *}^{-(i, j)} + V H_{w} }
			\end{array}
			\label{eqn:background-word-prob}
		\end{equation}
		
		Equation \ref{eqn:background-word-prob} is nearly identical to \ref{eqn:source-word-prob}, with the exception of the first term, the word-topic probability term, where $c_{z_{i, j}, *, T_d, *}^{-(i, j)}$ refers to the count of words associated with word-topic $z_{i, j}$ in document-type $T_d$, not considering the current word. The second term, the word probability term, is identical.

	\end{document}